\documentclass{article}

\usepackage[final]{corl_2026}

\makeatletter
\renewcommand{\@conferencelocation}{Austin, Texas, USA}
\makeatother

\usepackage[utf8]{inputenc}
\usepackage[T1]{fontenc}
\usepackage{url}
\usepackage{booktabs}
\usepackage{amsfonts}
\usepackage{amssymb}
\usepackage{amsmath}
\usepackage{bm}
\usepackage{nicefrac}
\usepackage{microtype}
\usepackage{graphicx}
\usepackage{multirow}
\usepackage{makecell}
\usepackage{wrapfig}
\usepackage{cleveref}

\renewcommand{\vec}[1]{\bm{#1}}
\newcommand{\mat}[1]{\mathbf{#1}}

\title{Robot-Factored World Models via Robot Rendering}

\author{%
  Byungjun Kim$^{1}$ \quad
  Taeksoo Kim$^{1}$ \quad
  Hyunsoo Cha$^{1}$ \quad
  Hanbyul Joo$^{1,2}$ \\
  $^{1}$Seoul National University \quad
  $^{2}$RLWRLD \\
  \texttt{\{byungjun.kim,taeksu98,243stephen,hbjoo\}@snu.ac.kr} \\
  \url{https://bjkim95.github.io/rofacto/}
}

\hypersetup{
  pdftitle={Robot-Factored World Models via Robot Rendering},
  pdfauthor={Byungjun Kim, Taeksoo Kim, Hyunsoo Cha, Hanbyul Joo}
}

\begin{document}

\maketitle

\begin{abstract}
Action-conditioned video world models predict future observations from an initial observation and an action signal.
We propose robot-factored world models (RoFacto), which move two robot-specific factors outside the world model to separate action realization from scene-response prediction.
First, action realization: each command sequence is rolled through the robot's own controller and kinematics without scene contact into a deployment-available nominal trajectory that does not require future realized states.
Second, robot rendering: this nominal trajectory is rendered through the robot URDF, factoring the robot's geometry, kinematics, and appearance out of the model and into explicit rendered robot geometry.
Paired end-effector and scene depth provide geometric cues for contact and occlusion, while camera-aware static RGB context and rendered robot geometry form a shared visual interface across viewpoints and embodiments through which the model learns scene responses.
Our experiments show that the rendered interface outperforms vector-conditioned baselines, supports transfer to unseen robot embodiments, and generates robot manipulation videos from retargeted human demonstrations.

\end{abstract}


\section{Introduction}
World models predict how an environment evolves in response to an agent's actions, so that behavior can be anticipated without acting in the real world~\cite{worldmodels}. Recent advances in video generative models have enabled world models that predict future observations directly in pixel space~\cite{nwm,irasim,ctrlworld,dreamdojo}. For robotics, this suggests an appealing use: conditioned on a candidate action, a video world model can roll out the future that action would produce, serving as a learned environment for policy improvement~\cite{worldgymnast} or for ranking behaviors before real execution~\cite{worldgym}. This raises a central question: how should robot actions be presented to a video world model?

For a physical robot, actions affect future observations through an embodiment-specific realization process before any scene interaction occurs. Consequently, a world model conditioned directly on robot-specific control signals must learn both how those signals are realized into robot motion and how the scene responds to that motion. Existing robot video world models largely keep this burden inside the network: they inject robot control signals through conditioning layers~\cite{irasim,ctrlworld,dreamdojo}, so the model-facing signal remains tied to an embodiment-specific action representation. Visual Action Prompts (VAP)~\cite{vap} suggest representing actions as image-space visual conditions. For robot world models, however, the key question is which robot signal should be rendered. VAP instantiates this idea by rendering logged robot state trajectories, which are visually aligned with the future video. These logged states can encode contact, compliance, latency, and closed-loop corrections, but are unavailable before execution. Rendering them may therefore expose interaction outcomes that the world model is meant to predict. The challenge is to identify a deployment-available visual conditioning signal that preserves scene-response prediction as the model's task.

To address this challenge, we introduce robot-factored world models (RoFacto), which condition video prediction on deployment-available nominal trajectories through a shared visual interface. Given an action sequence, we roll it through the robot's own controller and kinematics without scene contact, producing this nominal trajectory. It approximates free-space motion while leaving contact-mediated deviations to be predicted, subject to the fidelity of the robot-local controller model.
We render the nominal trajectory through the robot URDF into camera-aligned robot geometry. Together, camera-aware static RGB/depth context and rendered robot geometry form a shared visual world-model interface for fixed and dynamic viewpoints. The model sees the action as visible robot geometry and learns how the scene responds around it. This rendered robot geometry localizes the action in image space, but RGB alone cannot resolve the robot's spatial relationship to the scene. 
We therefore augment the interface with end-effector and scene depth, providing geometric cues for proximity, contact, and occlusion beyond image-plane overlap.

Because the interface represents actions as rendered robot geometry rather than embodiment-specific control signals, it decouples world-model conditioning from robot-specific action representations.
This supports embodiment generalization to unseen robots and grounded robot-interaction video generation from retargeted human demonstrations.

In summary, our contributions are as follows:

(1) We introduce a deployment-realistic visual world-model interface that realizes actions into nominal robot trajectories rendered as camera-aligned URDF mesh RGB and end-effector depth, with camera-aware static RGB/depth context factoring out scene appearance and viewpoint.

(2) We make the interface depth-aware by augmenting it with end-effector and scene depth, providing geometric cues for proximity, occlusion, and likely contact beyond RGB image-plane overlap.

(3) We show that the rendered interface is shared across robot embodiments and motion sources, supporting transfer to unseen robots and robot-interaction video generation from retargeted human demonstrations.

\section{Related Work}

\paragraph{Action-Conditioned Robot World Models.}

World models predict how observations evolve under actions and have been used for imagination, planning, policy evaluation, and policy improvement~\cite{worldmodels,planet,dreamerv3,daydreamer,td-mpc2}. Recent video world models extend this idea to pixel-space rollouts for games, driving, navigation, and robot manipulation~\cite{cosmos,dreamgen,worldgym,worldgymnast,irasim,ctrlworld,dreamdojo,genie,diamond,gamengen,unisim,ivideogpt,aether,seva,gaia,nwm}. These approaches differ primarily in how actions are represented and presented to the model. Some models condition directly on robot-specific action signals~\cite{irasim,ctrlworld,worldgym}, while others learn latent or abstract action representations from large-scale video~\cite{dreamdojo,adaworld,coworld,clam,univla,lawm}. Direct action interfaces remain tied to embodiment-specific control spaces, whereas latent-action approaches hide embodiment geometry behind learned abstractions. Our work instead conditions on nominal trajectories computed without scene contact, exposing embodiment geometry without requiring future realized robot states.

\paragraph{Visual Conditioning for Video Generation.}

Visual conditioning is a standard mechanism for steering image and video diffusion models with spatial signals such as depth, pose, edges, masks, optical flow, trajectories, and reference frames~\cite{controlnet, controlvideo, videocomposer, animatediff, motionctrl, dragnuwa, tora, mofa, trackgo, vace}. 
A parallel literature treats object motion, camera motion, drag edits, and point tracks as direct video conditions~\cite{motion-prompting, motionstream, freetraj, trailblazer, peekaboo, image-conductor}. 
Together, these works establish visual conditioning as a general mechanism for controlling image and video generation.

For embodied video prediction, visual controls can represent the acting body itself. Visual Action Prompts~\cite{vap} render robot states into image-space prompts, while Dexterous World Models~\cite{dwm} condition video generation on static scene renderings and hand mesh trajectories to model interaction-induced scene dynamics. We share the visual-conditioning principle of these approaches but differ in which robot signal is rendered: the nominal trajectory computed before scene interaction, rather than logged states that already reflect interaction outcomes.

\section{Method}

\subsection{Robot-Factored Visual World-Model Interface}
\label{sec:method:overview}

An action-conditioned robot world model predicts a future video $\mat{V}_{1:F}$ from the current observation and a proposed action sequence $\vec{a}_{1:F}$. 
Conditioning directly on $\vec{a}_{1:F}$ asks the video model to learn two distinct processes: how robot actions are realized into robot motion, and how the scene responds to that motion. 
The former depends on robot-specific embodiment and control, whereas the latter is the shared prediction problem that a world model should solve. We therefore separate action realization from scene-response prediction and model the realization process explicitly before world-model conditioning.

We represent this realization process with an operator $\Phi_R$ that maps actions into a nominal trajectory:
\begin{equation}
\vec{q}_{1:F} = \Phi_R(\vec{a}_{1:F}; \vec{q}_0),
\label{eq:nominal-realization}
\end{equation}
where $\vec{q}_{1:F}$ is the nominal trajectory realized from the action prior to scene interaction.
We then render this nominal trajectory along the target camera trajectory $\mathcal{C}_{1:F}$:
\begin{equation}
(\mat{M}^{\mathrm{rgb}}_{1:F}, \mat{D}^{\mathrm{eef}}_{1:F}) = \Pi_R(\vec{q}_{1:F}; \mathcal{C}_{1:F}),
\label{eq:robot-rendering}
\end{equation}
where $\mat{M}^{\mathrm{rgb}}_{1:F}$ is the robot mesh RGB video and $\mat{D}^{\mathrm{eef}}_{1:F}$ is end-effector-only depth. The rendering operator $\Pi_R$ uses the robot URDF and geometry to project the nominal trajectory into the target camera frame.
In parallel, the initial scene state $\mat{S}_0$ is represented through a static context stream rendered along the same camera trajectory:
\begin{equation}
(\mat{B}^{\mathrm{rgb}}_{1:F},
 \mat{D}^{\mathrm{scene}}_{1:F})
=
\Pi_S(\mat{S}_0; \mathcal{C}_{1:F}),
\label{eq:static-context}
\end{equation}
where $\mat{B}^{\mathrm{rgb}}_{1:F}$ and $\mat{D}^{\mathrm{scene}}_{1:F}$ provide appearance and geometric context in the target camera coordinates.
Together, these streams define the conditioning interface of the world model, which learns
\begin{equation}
p_{\theta}
\left(
\mat{V}_{1:F}
\mid
\mat{B}^{\mathrm{rgb}}_{1:F},
\mat{D}^{\mathrm{scene}}_{1:F},
\mat{M}^{\mathrm{rgb}}_{1:F},
\mat{D}^{\mathrm{eef}}_{1:F},
\mathcal{T}
\right),
\label{eq:method:our-formulation}
\end{equation}
where $\mathcal{T}$ is the text prompt provided to the video model. It contains scene context only and excludes descriptions of the intended action or future outcome.

This decomposition turns robot-specific control signals into model-facing visible robot geometry.
The realization and rendering operators are fixed preprocessing steps, leaving the learned model with the shared problem of predicting scene response around the rendered robot motion.

\begin{figure}[t]
  \centering
  \includegraphics[width=\linewidth]{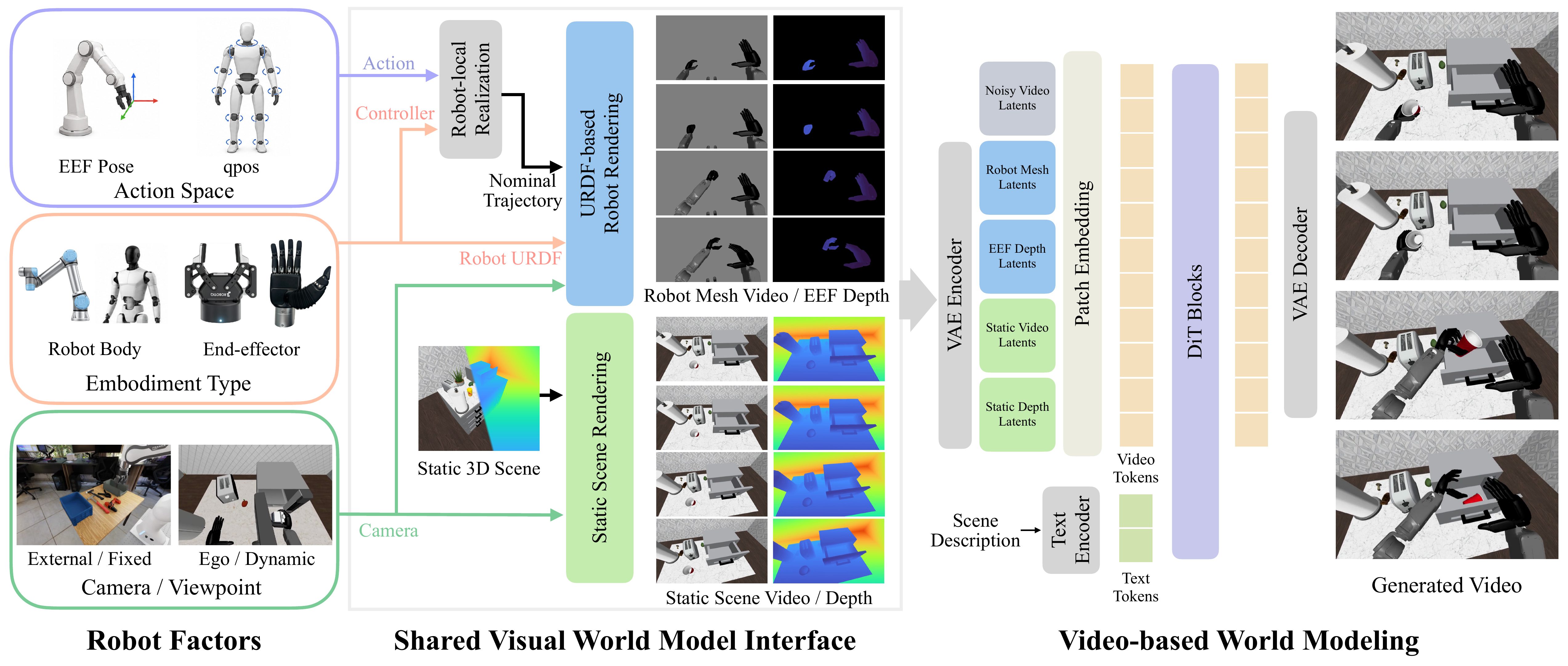}
  \caption{\textbf{Visual world-model interface.} Static context carries scene and viewpoint; rendered nominal robot geometry carries action; the diffusion model predicts scene response.}
  \label{fig:architecture}
\end{figure}

\subsection{Nominal Trajectory Conditioning}
\label{sec:method:nominalization}

\begin{figure}[t]
  \centering
  \includegraphics[width=\linewidth]{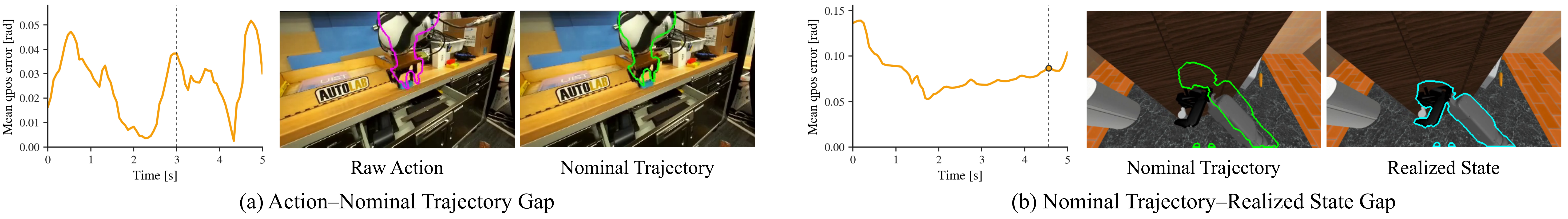}
  \caption{
\textbf{Action-to-state realization gaps.}
(a) Robot-specific controllers and hardware constraints create a gap between raw actions and nominal trajectories. (b) Scene interaction creates a gap between nominal trajectories and realized states. The nominal trajectory serves as the deployment-available conditioning signal.
}
  \label{fig:action-nominal-realized}
\end{figure}

A deployment-time visual conditioning signal must satisfy two requirements: it must be available at inference time and remain informative about the robot motion realized from the action.
The \emph{raw action} $\vec{a}_{1:F}$ is what a policy, teleoperator, or dataset controller emits, such as joint-space or task-space control commands. It is available at deployment, but its meaning is tied to a particular robot and controller.
The \emph{nominal trajectory} $\vec{q}_{1:F}$ lies between action and interaction: it is the robot-only motion realized by the robot's own controller and kinematics without scene contact. For a specified action sequence, it can be computed without observing future interaction outcomes.
The \emph{realized state} $\vec{x}_{1:F}$ is the robot state logged after executing the action in a scene. It is visually aligned with the future video, but it can encode scene-interaction outcomes and is unavailable before execution. Rendering realized states as prompts serves as an oracle diagnostic; the deployment-available conditioning signal is the nominal trajectory.
Figure~\ref{fig:action-nominal-realized} illustrates why the nominal trajectory serves as the appropriate middle signal for world-model conditioning.

The mismatch between these signals can be decomposed into two gaps. The \emph{action-realization gap} is the difference between the raw action and the controller-realized nominal motion. 
The \emph{nominal-realized gap} is the difference between the nominal trajectory and the state actually observed in a contact-rich rollout.
Our factorization assigns the first gap to the robot-specific realization process and leaves the second gap, together with object motion and occlusion changes, to the world model. 
Each robot instantiates $\Phi_R$ with its own controller, so robot-specific action semantics are resolved before the video model receives its visual conditioning streams.

\subsection{Camera-Aware Static Context}
\label{sec:method:static-context}

The static context stream, defined in \cref{eq:static-context}, provides the world model with a camera-aligned representation of the initial scene state.
It supplies scene appearance, scene depth, and viewpoint information independent of robot interaction, so the model can focus on interaction-induced change while the static stream preserves scene content~\cite{dwm}.
    
When a static scene representation and camera trajectory are available, $\Pi_S$ produces static scene RGB and depth along the same $\mathcal{C}_{1:F}$ used by the target video.
For fixed cameras, we instantiate the same stream by repeating the initial frame, including any visible robot, across the prediction horizon.
This formulation gives fixed- and dynamic-view clips a common camera-aligned conditioning interface, while action information enters separately through the rendered nominal robot geometry.

\subsection{Rendered Robot Action Prompts}
\label{sec:method:visual-prompt}
The renderer $\Pi_R$ in \cref{eq:robot-rendering} turns a nominal trajectory into rendered robot geometry in the target camera frame.
We render the robot as URDF mesh RGB, preserving link geometry, wrist offsets, gripper shape, and end-effector structure.
This makes the action visible to the video model in the same visual coordinates as the target video.

RGB mesh rendering localizes the robot in image space, and depth provides cues about whether an end-effector overlapping an object is in front of it, behind it, or at contact depth.
We therefore include end-effector-only depth $\mat{D}^{\mathrm{eef}}_{1:F}$ alongside static scene depth $\mat{D}^{\mathrm{scene}}_{1:F}$.
We focus depth rendering on the end-effector, where interaction is most likely to occur.
Together, these depth signals help disambiguate proximity, likely contact, and occlusion beyond apparent 2D overlap.

\subsection{World Model Training}
\label{sec:method:world-model}

We instantiate Eq.~\ref{eq:method:our-formulation} with a latent video diffusion model. Following the residual-dynamics formulation of Dexterous World Models~\cite{dwm}, we use a video inpainting backbone~\cite{videox-fun}, where the static context video serves as the conditioning input and a full mask is applied over the prediction horizon. The model therefore learns to generate interaction-induced scene changes conditioned on the static scene context and rendered robot geometry.

All conditioning streams are encoded by the pretrained video VAE~\cite{vae} of the backbone model~\cite{wan2025}. The static RGB context, rendered robot RGB, scene depth, and end-effector depth streams are encoded and concatenated with the noisy video latent:
\begin{equation}
\vec{c}
=
\left[
\mathcal{E}(\mat{B}^{\mathrm{rgb}}_{1:F}),
\mathcal{E}(\mat{M}^{\mathrm{rgb}}_{1:F}),
\mathcal{E}(\mat{D}^{\mathrm{scene}}_{1:F}),
\mathcal{E}(\mat{D}^{\mathrm{eef}}_{1:F})
\right].
\end{equation}
Here $\mathcal{E}$ denotes the pretrained video VAE encoder applied to each conditioning stream.

Training minimizes the latent flow-matching objective~\cite{lipman2023flowmatching,liu2023flow}. For a sampled noise latent $\vec{\epsilon}$ and noise level $\sigma$, we form $\vec{z}_{\sigma}=(1-\sigma)\vec{z}_0+\sigma\vec{\epsilon}$ and use the target velocity $\vec{u}=\vec{\epsilon}-\vec{z}_0$:

\begin{equation}
\mathcal{L}
=
\mathbb{E}_{\vec{z}_0,\sigma,\vec{\epsilon}}
\left[
\left\|
\vec{u}
-
v_\theta
\left(
\vec{z}_{\sigma},
\sigma
\mid
\vec{c},
\mathcal{T}
\right)
\right\|_2^2
\right].
\end{equation}

The text prompt $\mathcal{T}$ contains scene context only and excludes descriptions of the intended action or future outcome.

\section{Experiments}

\subsection{Experimental Setup}
\label{sec:exp:setup}

\paragraph{Datasets.}
We use the enhanced-extrinsic subset of DROID~\cite{droid} and RoboCasa-GR1 rollouts collected in RoboCasa~\cite{robocasa} by executing the humanoid VLA DiT4DiT~\cite{dit4dit} across 24 tasks with 50 episodes per task.
Both datasets are processed into 81-frame, $480{\times}832$, 16~fps clips, yielding 41,642 DROID clips and 9,380 RoboCasa-GR1 clips.
Evaluation uses 256 held-out DROID clips and 128 held-out RoboCasa-GR1 clips.
Additional studies use RoboTwin~2.0~\cite{robotwin} and HRDexDB~\cite{hrdexdb} for embodiment transfer, RoboLab~\cite{robolab} for task-outcome prediction, and DexYCB~\cite{dexycb} for human-to-robot video generation.

\paragraph{Nominalization.}
For DROID, the raw teleoperator joint and gripper targets are replayed in a scene-free, robot-only Isaac Lab~\cite{mittal2025isaaclab} environment, so the robot's own controller and actuation limits yield a physically trackable nominal trajectory.
For RoboCasa-GR1, the DiT4DiT~\cite{dit4dit} controller actions are replayed from the clip-start state in a collision-free shadow rollout to obtain the Fourier GR-1 nominal state.
In both cases, the nominal trajectory is available at inference and rendered as the robot conditioning signal.

\paragraph{Baselines.}
We compare against \emph{Ctrl-World}~\cite{ctrlworld}, a 7-DoF Cartesian pose-conditioned world model built on Stable Video Diffusion (SVD)~\cite{svd}, and an \emph{AdaLN}~\cite{peebles2023scalable} \emph{state vector} baseline that feeds the same deployment-available nominal trajectory as numeric state values.
For the SVD-based comparison, we retrain the same backbone with pose conditioning or the rendered interface and evaluate on the DROID held-out set.
For Wan~\cite{wan2025}/Wan2.1-Fun InP~\cite{videox-fun}, we compare numeric state-vector conditioning against our rendered interface with mesh and depth.
Both Wan variants use rank-64 LoRA adaptation of the same pretrained backbone.
The joint AdaLN baseline uses a zero-padded 46D union of the 7D DROID pose--gripper state and 39D GR-1 state, stacked over four frames into 184D.
Implementation and evaluation details appear in Appendix~\ref{sec:supp:training}.

\subsection{Main Comparison}
\label{sec:exp:main}

\begin{table}[t]
\centering
\small
\setlength{\tabcolsep}{3pt}
\caption{\textbf{Main action-interface comparison.}
  Each backbone cell lists the model, its training resolution, and its training data (D+R: DROID+RoboCasa-GR1).
  SVD 1.5B trains on DROID only and is therefore not evaluated on RoboCasa-GR1 (dashes); since lower resolution inflates reconstruction metrics, numbers are comparable only within a backbone.
  DROID and RoboCasa-GR1 are reported separately rather than averaged; best per backbone in \textbf{bold}.}
\label{tab:main-results}
\begin{tabular}{l l ccc ccc}
\toprule
\multirow{2}{*}{Backbone} & \multirow{2}{*}{Method}
  & \multicolumn{3}{c}{DROID} & \multicolumn{3}{c}{RoboCasa-GR1} \\
\cmidrule(lr){3-5}\cmidrule(lr){6-8}
& & PSNR$\uparrow$ & SSIM$\uparrow$ & LPIPS$\downarrow$
   & PSNR$\uparrow$ & SSIM$\uparrow$ & LPIPS$\downarrow$ \\
\midrule
\multirow{2}{*}{\makecell[l]{SVD 1.5B\\[1pt]\scriptsize $192{\times}320$, DROID}}
  & Ctrl-World pose~\cite{ctrlworld}
    & 23.15 & 0.884 & 0.104 & -- & -- & -- \\
  & Rendered mesh (ours)
    & \textbf{25.05} & \textbf{0.899} & \textbf{0.091} & -- & -- & -- \\
\midrule
\multirow{2}{*}{\makecell[l]{Wan 2.1 14B\\[1pt]\scriptsize $480{\times}832$, D+R}}
  & AdaLN state vector
    & 18.57 & 0.824 & 0.224 & 17.67 & 0.835 & 0.194 \\
  & \makecell[l]{Rendered mesh $+$\\ EEF/scene depth (ours)}
    & \textbf{21.87} & \textbf{0.859} & \textbf{0.178} & \textbf{24.61} & \textbf{0.889} & \textbf{0.131} \\
\bottomrule
\end{tabular}
\end{table}

\paragraph{Video reconstruction.}
Table~\ref{tab:main-results} compares action interfaces using clip-averaged PSNR, SSIM, and LPIPS~\cite{lpips}.
Within the backbone-controlled SVD comparison, the rendered interface outperforms the Ctrl-World-style Cartesian pose condition.
Within the jointly trained Wan setting, the rendered interface outperforms the AdaLN state-vector baseline on both DROID and RoboCasa-GR1.
These results support the central interface design: rendered robot geometry gives the video model direct pixel-space evidence of robot motion, while numeric action or state conditions require the model to learn an additional grounding map.
Because the backbone cells differ in resolution and training data, we compare only within each backbone.

The advantage also persists outside the ground-truth robot region: on all 256 DROID clips, PSNR is 23.71 versus 20.87, and LPIPS is 0.150 versus 0.176 for Ours and AdaLN, respectively.
The same trend holds on the 82 RoboCasa-GR1 clips with available robot masks (Appendix~\ref{sec:supp:masked}).
This shows that the gain is not confined to robot pixels, although these scores are not a direct measure of object dynamics.

\begin{table}[t]
\centering
\small
\caption{\textbf{Task-outcome prediction accuracy (\%).}
Both sets are balanced between success and failure (chance: 50\%).
RoboCasa-GR1 uses clips whose outcomes the judge correctly identifies from ground-truth video; RoboLab includes 20 settling frames.}
\label{tab:outcomes}
\begin{tabular}{lcc}
\toprule
Method & RoboCasa-GR1 ($n{=}160$) & RoboLab ($n{=}100$) \\
\midrule
AdaLN & 70.6 & 49.0 \\
Ours & \textbf{75.0} & \textbf{66.0} \\
\bottomrule
\end{tabular}
\end{table}

\paragraph{Task-outcome prediction.}
\label{sec:exp:outcomes}
To probe task-relevant information beyond reconstruction, an automatic video judge predicts success or failure from each generated rollout, scored against simulator predicates.
On RoboCasa-GR1, we use 160 clips balanced by outcome and matched by task, restricted to outcomes observable in the ground-truth video within the prediction window.
On RoboLab, a simulation of the DROID setup, we evaluate 100 policy rollouts over five tasks, with ten successes and ten failures per task.
Table~\ref{tab:outcomes} reports classification accuracy, not robot task success rate.
Ours is numerically higher in both settings, with a small gap on RoboCasa-GR1 and a larger gap on RoboLab.
These results assess task-outcome information in generated videos conditioned on recorded action sequences.
Appendix~\ref{sec:supp:outcomes} gives the selection protocol, full-set RoboCasa-GR1 results, and confusion counts documenting a bias toward predicting failure, especially for AdaLN on RoboLab.

\paragraph{Qualitative comparison.}

\begin{figure}[t]
  \centering
  \includegraphics[width=\linewidth]{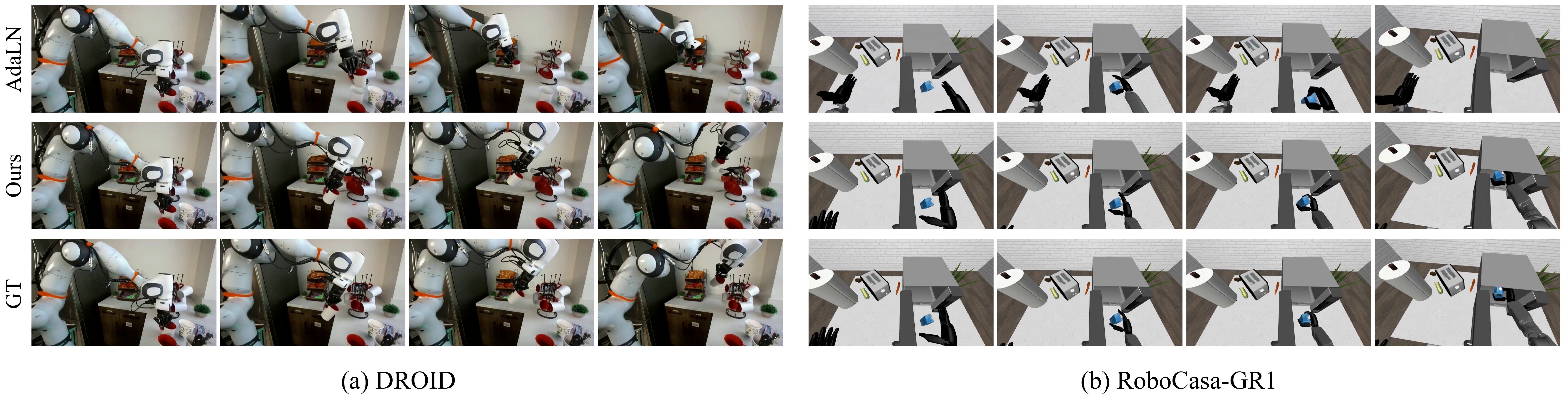}
  \caption{\textbf{Qualitative comparison.} Rendered robot geometry localizes robot-driven scene changes.}
  \label{fig:qualitative-comparison}
\end{figure}

Figure~\ref{fig:qualitative-comparison} emphasizes cases where reconstruction metrics are least diagnostic: small object motion, brief contact, occlusion ordering, missed grasps, and no-contact outcomes.
The AdaLN state-vector baseline often preserves scene appearance, but the intended action is weak or misplaced.
In contrast, rendered robot geometry exposes the action directly in the target camera frame, so the predicted scene changes are better localized around the robot motion and contact region.

\subsection{Ablation Study}
\label{sec:exp:ablations}

Table~\ref{tab:ablations} isolates the two factors introduced in Sections~\ref{sec:method:nominalization} and~\ref{sec:method:visual-prompt}: nominal trajectory conditioning and the depth-aware rendered interface.
All variants in this table are trained and evaluated on DROID, keeping the dataset and backbone fixed.

\begin{wraptable}{r}{0.48\textwidth}
\centering
\vspace{-\baselineskip}
\footnotesize
\setlength{\tabcolsep}{3pt}
\caption{\textbf{Ablation study on DROID.}}
\label{tab:ablations}
\begin{tabular}{lccc}
\toprule
Variant & PSNR $\uparrow$ & SSIM $\uparrow$ & LPIPS $\downarrow$ \\
\midrule
Raw action mesh               & 21.34 & 0.848 & 0.187 \\
Nominal mesh                  & 21.76 & 0.858 & 0.179 \\
\makecell[l]{Nominal mesh $+$\\ EEF/scene depth} & 21.96 & 0.856 & 0.179 \\
\bottomrule
\end{tabular}
\end{wraptable}

\paragraph{Nominal trajectory conditioning.}
The first two rows compare raw-action mesh against nominal trajectory rendering, conditioning on mesh RGB only in both cases.
This isolates the action-realization gap: the raw-action mesh exposes the command before robot-local realization, whereas nominal trajectory rendering shows the controller-realized robot motion available at deployment.
Nominal trajectory rendering improves over the raw-action mesh across metrics (Table~\ref{tab:ablations}), showing that assigning action realization to the robot-local stack produces better aligned rendered robot geometry from deployment-available motion.

\begin{table}[b]
\centering
\small
\caption{\textbf{Training and inference state conditioning.}
Logged/logged is an oracle setting. Models are trained separately per dataset.}
\label{tab:state-conditioning}
\begin{tabular}{lllccc}
\toprule
Dataset & Train state & Inference state & PSNR $\uparrow$ & SSIM $\uparrow$ & LPIPS $\downarrow$ \\
\midrule
DROID & nominal & nominal & 21.74 & 0.856 & 0.179 \\
DROID & logged & logged & 22.37 & 0.861 & 0.174 \\
DROID & logged & nominal & 21.12 & 0.846 & 0.187 \\
\midrule
RoboCasa-GR1 & nominal & nominal & 25.70 & 0.904 & 0.120 \\
RoboCasa-GR1 & logged & logged & 28.26 & 0.918 & 0.112 \\
RoboCasa-GR1 & logged & nominal & 24.69 & 0.897 & 0.126 \\
\bottomrule
\end{tabular}
\end{table}

\paragraph{Training and inference state conditioning.}
Table~\ref{tab:state-conditioning} compares nominal and logged realized-state prompts at training and inference, using mesh RGB without depth.
Logged/logged obtains higher reconstruction scores, but requires future realized states unavailable before execution.
Replacing logged prompts with nominal prompts at inference degrades the same logged-trained model on both datasets, whereas nominal/nominal performs better than logged/nominal.

\begin{figure}[t]
  \centering
  \includegraphics[width=\linewidth]{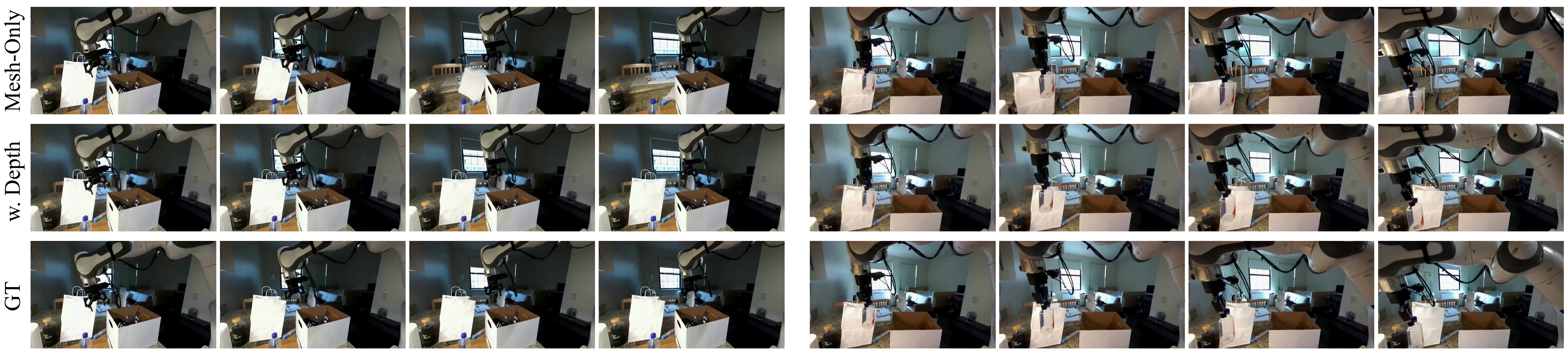}
  \caption{\textbf{Depth ablation.}
  End-effector and scene depth help distinguish contact-relevant proximity from image-plane overlap.}
  \label{fig:depth-ablation}
\end{figure}

\paragraph{Depth-aware rendered interface.}
The final row of Table~\ref{tab:ablations} adds end-effector and scene depth on top of nominal trajectory rendering.
This evaluates the depth-aware interface from Section~\ref{sec:method:visual-prompt}: RGB mesh rendering localizes robot motion in the image plane, while the depth pair exposes contact-relevant 3D proximity and occlusion order.
Adding depth further improves PSNR, with slightly lower SSIM and similar LPIPS (Table~\ref{tab:ablations}).
While these changes in full-frame reconstruction are modest, Figure~\ref{fig:depth-ablation} highlights qualitative differences in contact-ambiguous cases: without depth, the model can mistake image-plane overlap for contact with a more distant object.

\subsection{Transfer Across Embodiments and Motion Sources}
\label{sec:exp:edits}

\begin{table}[t]
\centering
\small
\setlength{\tabcolsep}{5pt}
\caption{\textbf{Embodiment transfer.} Full-frame reconstruction metrics.
(A) Both methods transfer zero-shot, with 40 clips per embodiment.
(B) HRDexDB uses 84 held-out clips; target clips denotes adaptation data.
ZS: zero-shot; FT: fine-tuned.}
\label{tab:transfer}
\begin{tabular}{llcccc}
\toprule
Embodiment & Method & Target clips & PSNR$\uparrow$ & SSIM$\uparrow$ & LPIPS$\downarrow$ \\
\midrule
\multicolumn{6}{l}{\textit{A. RoboTwin: compatible pose--gripper interface}} \\
ur5-wsg & AdaLN (ZS) & 0 & 22.96 & 0.946 & 0.176 \\
 & Ours (ZS) & 0 & \textbf{26.94} & \textbf{0.962} & \textbf{0.141} \\
ARX-X5 & AdaLN (ZS) & 0 & 17.33 & 0.930 & 0.200 \\
 & Ours (ZS) & 0 & \textbf{22.77} & \textbf{0.965} & \textbf{0.136} \\
\midrule
\multicolumn{6}{l}{\textit{B. HRDexDB}} \\
xArm 6--Inspire F1 & Ours (ZS) & 0 & \underline{21.00} & \underline{0.857} & \underline{0.201} \\
 & AdaLN (FT) & 575 & 19.29 & 0.837 & 0.221 \\
 & Ours (FT) & 575 & \textbf{23.01} & \textbf{0.885} & \textbf{0.164} \\
\bottomrule
\end{tabular}
\end{table}

\paragraph{Compatible single-arm interfaces.}
On RoboTwin~2.0, we evaluate two unseen arm--gripper combinations, ur5-wsg and ARX-X5, on eight tasks with five episodes each.
Both methods are applied zero-shot to the same nominal motions: AdaLN receives end-effector pose and gripper state through its existing input, while Ours receives the rendered geometry.
Ours achieves higher reconstruction quality on both embodiments (Table~\ref{tab:transfer}A).
Thus, a compatible vector interface can also be applied to a new robot without retraining; the question is the quality of that transfer, not whether it is possible.
Qualitative examples appear in Appendix~\ref{sec:supp:embodiments}.

\begin{figure}[t]
  \centering
  \includegraphics[width=\linewidth]{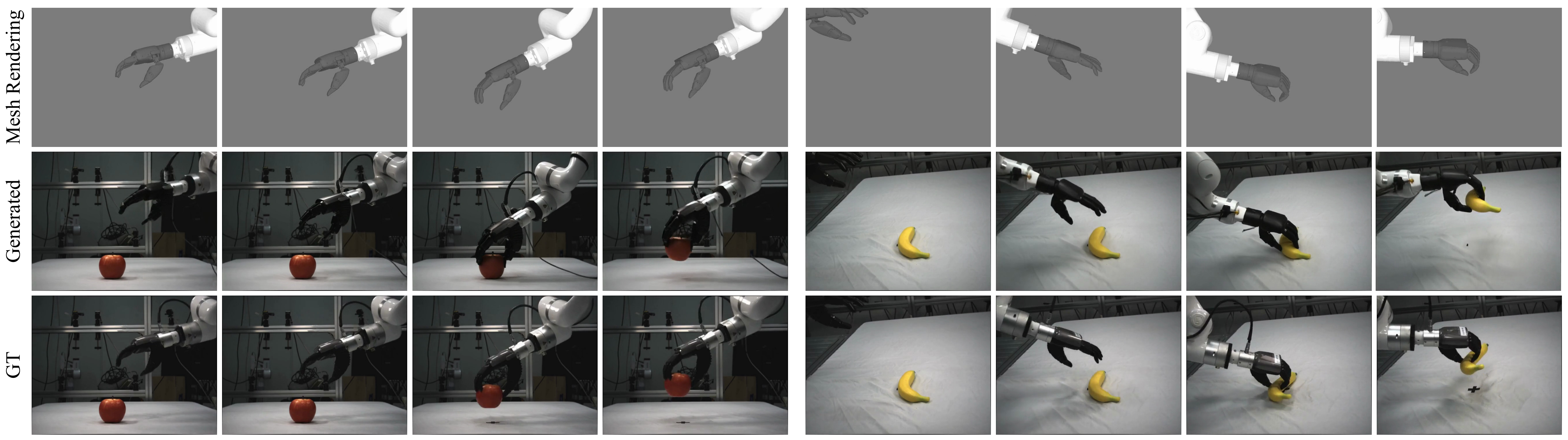}
  \caption{\textbf{Zero-shot embodiment transfer.}
  HRDexDB contains an unseen xArm 6--Inspire F1 pairing; rendered URDF motion still drives the predicted scene response.}
  \label{fig:embodiment-composition}
  \vspace{-2mm}
\end{figure}

\paragraph{Dexterous embodiment transfer.}
Figure~\ref{fig:embodiment-composition} illustrates zero-shot video prediction on HRDexDB with an unseen xArm 6--Inspire F1 embodiment.
Conditioned on rendered robot motion, the model generates grasping and lifting interactions without target-robot training.
For quantitative evaluation, we use 84 clips from held-out objects, comparing zero-shot Ours with both methods fine-tuned on the same 575 target-robot clips (Table~\ref{tab:transfer}B).
The rendered interface remains unchanged; our AdaLN adaptation uses an expanded pose-and-hand-state input with a reinitialized conditioning head.
Ours improves with fine-tuning and exceeds this adapted baseline under the matched training budget.
Table~\ref{tab:transfer} reports zero-shot transfer and target-robot adaptation separately; Appendix~\ref{sec:supp:transfer} details the protocols.

\begin{figure}[t]
  \centering
  \includegraphics[width=\linewidth]{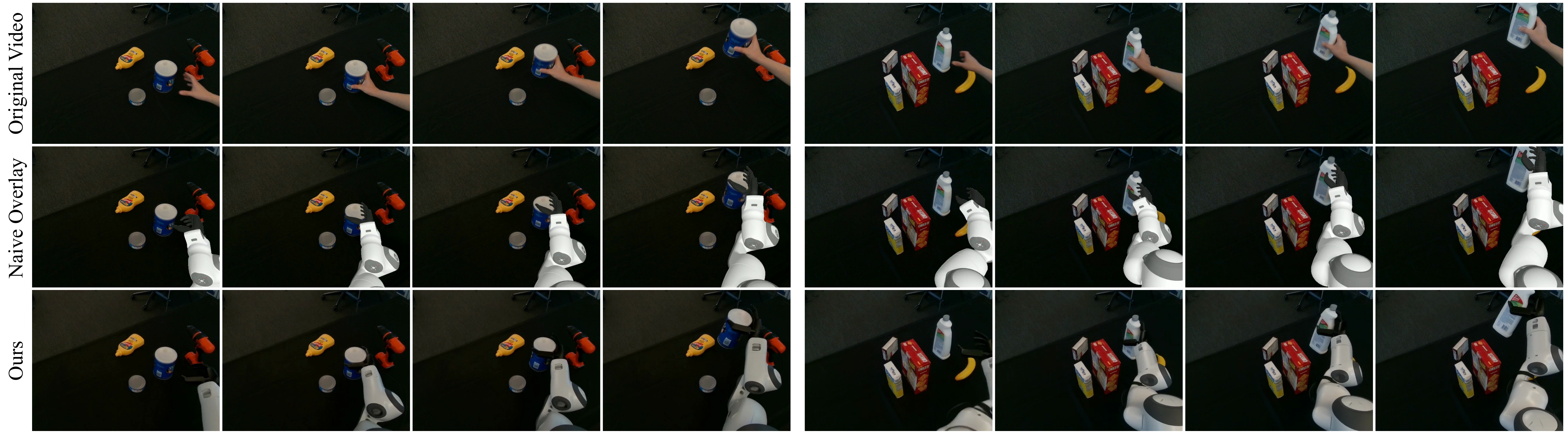}
  \caption{\textbf{Application: human demonstration to robot video.}
  Retargeted human motion is rendered as robot geometry and converted into a robot interaction rollout.}
  \label{fig:h2r-retargeting}
  \vspace{-2mm}
\end{figure}

\begin{samepage}
\paragraph{Human demonstration to robot video.}
\label{sec:exp:h2r}
As a downstream application of the shared rendered interface, Figure~\ref{fig:h2r-retargeting} takes DexYCB~\cite{dexycb} human manipulation videos, retargets the hand motion to a robot~\cite{qin2023anyteleop,kim2025pyroki}, and renders the result as the same mesh-and-depth interface used for robot data.
Because the motion originates from retargeted human demonstrations, this application illustrates that the rendered interface is decoupled from controller-specific action representations: the world model consumes rendered robot geometry regardless of the motion source.
\par
\end{samepage}

\section{Conclusion and Limitations}

We present robot-factored world models, which factor action realization and robot rendering out of the video world model.
Actions are realized into deployment-available nominal trajectories and rendered as camera-aligned robot geometry, so the model receives the action as visible robot geometry and learns how the scene responds around it.
Experiments support the interface through reconstruction, embodiment transfer, and simulation task-outcome prediction, with retargeted human demonstrations illustrating a further motion source.

The interface requires a known URDF, camera-to-robot calibration, and a robot-local controller model.
Dynamic views additionally require a static scene representation and camera trajectory; real-world reconstruction may introduce appearance and geometry errors.
Nominal replay avoids future-state inputs for a specified action sequence, but does not remove outcome correlations already present in logged closed-loop actions.
Depth is a geometric cue, not a guarantee of correct contact physics, and robot-excluded reconstruction remains an imperfect measure of object dynamics.
Our task-outcome study uses simulator labels and an automatic video judge, not physical execution, action ranking, or policy improvement.
Limited failure data can make subtle differences between successful and failed interactions difficult to capture.
Incorporating more diverse failure examples is a promising direction for improving outcome prediction.

\clearpage
\acknowledgments{This work was supported by RLWRLD, Institute of Engineering Research at Seoul National University, KIST Institutional Program (26Z9020), NRF grant funded by the Korean government (MSIT) [No. RS-2025-25396144 and No. RS-2026-25540546], and IITP grant funded by the Korea government (MSIT) [No. RS-2024-00439854, No. RS-2025-25442338, and No. RS-2021-II211343]. This work was also supported by the National AI Project of the Ministry of Science and ICT and the AIIS. H. Joo is the corresponding author.
}

\bibliography{references}

@String{CVPR    = {CVPR}}

@String{ICCV    = {ICCV}}

@String{ECCV    = {ECCV}}

@String{NIPS    = {NeurIPS}}

@String{AAAI    = {AAAI}}

@String{ICLR    = {ICLR}}

@String{ICML    = {ICML}}

@String{RSS     = {RSS}}

@String{CORL    = {CoRL}}

@String{ICRA    = {ICRA}}

@String{IROS    = {IROS}}

@String{SIGGRAPH     = {ACM SIGGRAPH}}

@String{SIGGRAPHASIA = {ACM SIGGRAPH Asia}}

@article{robotwin,
  title={{RoboTwin 2.0}: A Scalable Data Generator and Benchmark with Strong Domain Randomization for Robust Bimanual Robotic Manipulation},
  author={Chen, Tianxing and Chen, Zanxin and Chen, Baijun and Cai, Zijian and Liu, Yibin and Li, Zixuan and Liang, Qiwei and Lin, Xianliang and Ge, Yiheng and Gu, Zhenyu and Deng, Weiliang and Guo, Yubin and Nian, Tian and Xie, Xuanbing and Chen, Qiangyu and Su, Kailun and Xu, Tianling and Liu, Guodong and Hu, Mengkang and Gao, Huan-ang and Wang, Kaixuan and Liang, Zhixuan and Qin, Yusen and Yang, Xiaokang and Luo, Ping and Mu, Yao},
  journal={arXiv preprint arXiv:2506.18088},
  year={2025}
}

@inproceedings{robolab,
  title={{RoboLab}: A High-Fidelity Simulation Benchmark for Analysis of Task Generalist Policies},
  author={Yang, Xuning and Dagli, Rishit and Zook, Alex and Hadfield, Hugo and Goyal, Ankit and Birchfield, Stan and Ramos, Fabio and Tremblay, Jonathan},
  booktitle=RSS,
  year={2026}
}

@article{dreamdojo,
  title={DreamDojo: A Generalist Robot World Model from Large-Scale Human Videos},
  author={Gao, Shenyuan and Liang, William and Zheng, Kaiyuan and Malik, Ayaan and Ye, Seonghyeon and Yu, Sihyun and Tseng, Wei-Cheng and Dong, Yuzhu and Mo, Kaichun and Lin, Chen-Hsuan and Ma, Qianli and Nah, Seungjun and Magne, Loic and Xiang, Jiannan and Xie, Yuqi and Zheng, Ruijie and Niu, Dantong and Tan, You Liang and Zentner, K.R. and Kurian, George and Indupuru, Suneel and Jannaty, Pooya and Gu, Jinwei and Zhang, Jun and Malik, Jitendra and Abbeel, Pieter and Liu, Ming-Yu and Zhu, Yuke and Jang, Joel and Fan, Linxi},
  journal={arXiv preprint arXiv:2602.06949},
  year={2026}
}

@article{dreamgen,
  title={DreamGen: Unlocking Generalization in Robot Learning through Video World Models},
  author={Jang, Joel and Ye, Seonghyeon and Lin, Zongyu and Xiang, Jiannan and Bjorck, Johan and Fang, Yu and Hu, Fengyuan and Huang, Spencer and Kundalia, Kaushil and Lin, Yen-Chen and others},
  journal={arXiv preprint arXiv:2505.12705},
  year={2025}
}

@article{cosmos,
  title={Cosmos world foundation model platform for physical AI},
  author={Agarwal, Niket and Ali, Arslan and Bala, Maciej and Balaji, Yogesh and Barker, Erik and Cai, Tiffany and Chattopadhyay, Prithvijit and Chen, Yongxin and Cui, Yin and Ding, Yifan and others},
  journal={arXiv preprint arXiv:2501.03575},
  year={2025}
}

@article{worldgym,
  title={WorldGym: World Model as An Environment for Policy Evaluation},
  author={Quevedo, Julian and Sharma, Ansh Kumar and Sun, Yixiang and Suryavanshi, Varad and Liang, Percy and Yang, Sherry},
  journal={arXiv preprint arXiv:2506.00613},
  year={2025}
}

@article{worldgymnast,
  title={World-Gymnast: Training Robots with Reinforcement Learning in a World Model},
  author={Sharma, Ansh Kumar and Sun, Yixiang and Lu, Ninghao and Zhang, Yunzhe and Liu, Jiarao and Yang, Sherry},
  journal={arXiv preprint arXiv:2602.02454},
  year={2026}
}

@inproceedings{worldmodels,
  title={Recurrent world models facilitate policy evolution},
  author={Ha, David and Schmidhuber, J{\"u}rgen},
  booktitle=NIPS,
  year={2018}
}

@inproceedings{vae,
  title={Auto-Encoding Variational Bayes},
  author={Kingma, Diederik P. and Welling, Max},
  booktitle=ICLR,
  year={2014}
}

@inproceedings{vap,
  title={Precise action-to-video generation through visual action prompts},
  author={Wang, Yuang and Wen, Chao and Guo, Haoyu and Peng, Sida and Qin, Minghan and Bao, Hujun and Zhou, Xiaowei and Hu, Ruizhen},
  booktitle=ICCV,
  year={2025}
}

@inproceedings{ctrlworld,
  title={Ctrl-World: A Controllable Generative World Model for Robot Manipulation},
  author={Guo, Yanjiang and Shi, Lucy Xiaoyang and Chen, Jianyu and Finn, Chelsea},
  booktitle=ICLR,
  year={2026}
}

@inproceedings{irasim,
  title={Irasim: A fine-grained world model for robot manipulation},
  author={Zhu, Fangqi and Wu, Hongtao and Guo, Song and Liu, Yuxiao and Cheang, Chilam and Kong, Tao},
  booktitle=ICCV,
  year={2025}
}

@inproceedings{robocasa,
  title={RoboCasa: Large-Scale Simulation of Everyday Tasks for Generalist Robots},
  author={Soroush Nasiriany and Abhiram Maddukuri and Lance Zhang and Adeet Parikh and Aaron Lo and Abhishek Joshi and Ajay Mandlekar and Yuke Zhu},
  booktitle=RSS,
  year={2024}
}

@inproceedings{droid,
  title={DROID: A Large-Scale In-The-Wild Robot Manipulation Dataset},
  author={Khazatsky, Alexander and Pertsch, Karl and Nair, Suraj and Balakrishna, Ashwin and Dasari, Sudeep and Karamcheti, Siddharth and Nasiriany, Soroush and Srirama, Mohan Kumar and Chen, Lawrence Yunliang and Ellis, Kirsty and others},
  booktitle=RSS,
  year={2024}
}

@article{mittal2025isaaclab,
  title={Isaac Lab: A GPU-Accelerated Simulation Framework for Multi-Modal Robot Learning},
  author={Mittal, Mayank and Roth, Pascal and Tigue, James and Richard, Antoine and Zhang, Octi and Du, Peter and Serrano-Mu{\~n}oz, Antonio and Yao, Xinjie and Z{\"u}rbr{\"u}gg, Ren{\'e} and Rudin, Nikita and others},
  journal={arXiv preprint arXiv:2511.04831},
  year={2025}
}

@inproceedings{wen2025stereo,
  title={FoundationStereo: Zero-Shot Stereo Matching},
  author={Wen, Bowen and Trepte, Matthew and Aribido, Joseph and Kautz, Jan and Gallo, Orazio and Birchfield, Stan},
  booktitle=CVPR,
  year={2025}
}

@article{gr00t,
  title={GR00T N1: An Open Foundation Model for Generalist Humanoid Robots},
  author={NVIDIA and Johan Bjorck and Fernando Casta{\~n}eda and Nikita Cherniadev and Xingye Da and Runyu Ding and Linxi Fan and Yu Fang and Dieter Fox and Fengyuan Hu and Spencer Huang and Joel Jang and Zhenyu Jiang and Jan Kautz and Kaushil Kundalia and Lawrence Lao and Zhiqi Li and Zongyu Lin and Kevin Lin and Guilin Liu and Edith Llontop and Loic Magne and Ajay Mandlekar and Avnish Narayan and Soroush Nasiriany and Scott Reed and You Liang Tan and Guanzhi Wang and Zu Wang and Jing Wang and Qi Wang and Jiannan Xiang and Yuqi Xie and Yinzhen Xu and Zhenjia Xu and Seonghyeon Ye and Zhiding Yu and Ao Zhang and Hao Zhang and Yizhou Zhao and Ruijie Zheng and Yuke Zhu},
  journal={arXiv preprint arXiv:2503.14734},
  year={2025}
}

@article{dit4dit,
  title={DiT4DiT: Jointly Modeling Video Dynamics and Actions for Generalizable Robot Control},
  author={Ma, Teli and Zheng, Jia and Wang, Zifan and Jiang, Chunli and Cui, Andy and Liang, Junwei and Yang, Shuo},
  journal={arXiv preprint arXiv:2603.10448},
  year={2026}
}

@misc{videox-fun,
  author={aigc-apps},
  title={VideoX-Fun: A Video Generation Pipeline for Diffusion Transformer},
  year={2026},
  publisher={GitHub},
  url={https://github.com/aigc-apps/VideoX-Fun}
}

@article{svd,
  title={Stable Video Diffusion: Scaling Latent Video Diffusion Models to Large Datasets},
  author={Blattmann, Andreas and Dockhorn, Tim and Kulal, Sumith and Mendelevitch, Daniel and Kilian, Maciej and Lorenz, Dominik and Levi, Yam and English, Zion and Voleti, Vikram and Letts, Adam and others},
  journal={arXiv preprint arXiv:2311.15127},
  year={2023}
}

@article{wan2025,
  title={Wan: Open and Advanced Large-Scale Video Generative Models},
  author={{Team Wan} and Wang, Ang and Ai, Baole and Wen, Bin and Mao, Chaojie and Xie, Chen-Wei and Chen, Di and Yu, Feiwu and Zhao, Haiming and Yang, Jianxiao and others},
  journal={arXiv preprint arXiv:2503.20314},
  year={2025}
}

@article{qwen3vl,
  title={Qwen3-VL Technical Report},
  author={Bai, Shuai and Cai, Yuxuan and Chen, Ruizhe and Chen, Keqin and Chen, Xionghui and others},
  journal={arXiv preprint arXiv:2511.21631},
  year={2025}
}

@inproceedings{hu2022lora,
  title={Lo{RA}: Low-Rank Adaptation of Large Language Models},
  author={Hu, Edward J. and Shen, Yelong and Wallis, Phillip and Allen-Zhu, Zeyuan and Li, Yuanzhi and Wang, Shean and Wang, Lu and Chen, Weizhu},
  booktitle=ICLR,
  year={2022}
}

@inproceedings{loshchilov2018decoupled,
  title={Decoupled Weight Decay Regularization},
  author={Loshchilov, Ilya and Hutter, Frank},
  booktitle=ICLR,
  year={2019}
}

@inproceedings{peebles2023scalable,
  title={Scalable Diffusion Models with Transformers},
  author={Peebles, William and Xie, Saining},
  booktitle=ICCV,
  year={2023}
}

@inproceedings{lipman2023flowmatching,
  title={Flow Matching for Generative Modeling},
  author={Lipman, Yaron and Chen, Ricky T. Q. and Ben-Hamu, Heli and Nickel, Maximilian and Le, Matthew},
  booktitle=ICLR,
  year={2023}
}

@inproceedings{liu2023flow,
  title={Flow Straight and Fast: Learning to Generate and Transfer Data with Rectified Flow},
  author={Liu, Xingchao and Gong, Chengyue and Liu, Qiang},
  booktitle=ICLR,
  year={2023}
}

@article{dwm,
  title={Dexterous World Models},
  author={Kim, Byungjun and Kim, Taeksoo and Lee, Junyoung and Joo, Hanbyul},
  journal={arXiv preprint arXiv:2512.17907},
  year={2025}
}

@article{hrdexdb,
  title={HRDexDB: A Large-Scale Dataset of Dexterous Human and Robotic Hand Grasps},
  author={Lim, Jongbin and Ha, Taeyun and Choi, Mingi and Kim, Jisoo and Kim, Byungjun and Jeon, Subin and Joo, Hanbyul},
  journal={arXiv preprint arXiv:2604.14944},
  year={2026}
}

@inproceedings{dexycb,
  title={{DexYCB}: A Benchmark for Capturing Hand Grasping of Objects},
  author={Chao, Yu-Wei and Yang, Wei and Xiang, Yu and Molchanov, Pavlo and Handa, Ankur and Tremblay, Jonathan and Narang, Yashraj S and Van Wyk, Karl and Iqbal, Umar and Birchfield, Stan and Kautz, Jan and Fox, Dieter},
  booktitle=CVPR,
  year={2021}
}

@inproceedings{jiang2025dexmimicgen,
  title={{DexMimicGen}: Automated Data Generation for Bimanual Dexterous Manipulation via Imitation Learning},
  author={Jiang, Zhenyu and Xie, Yuqi and Lin, Kevin and Xu, Zhenjia and Wan, Weikang and Mandlekar, Ajay and Fan, Linxi Jim and Zhu, Yuke},
  booktitle=ICRA,
  pages={16923--16930},
  year={2025}
}

@inproceedings{qin2023anyteleop,
  title={AnyTeleop: A General Vision-Based Dexterous Robot Arm-Hand Teleoperation System},
  author={Qin, Yuzhe and Yang, Wei and Huang, Binghao and Van Wyk, Karl and Su, Hao and Wang, Xiaolong and Chao, Yu-Wei and Fox, Dieter},
  booktitle=RSS,
  year={2023}
}

@inproceedings{kim2025pyroki,
  title={PyRoki: A Modular Toolkit for Robot Kinematic Optimization},
  author={Kim, Chung Min and Yi, Brent and Choi, Hongsuk and Ma, Yi and Goldberg, Ken and Kanazawa, Angjoo},
  booktitle=IROS,
  year={2025}
}

@article{curobo,
  title={cuRobo: Parallelized Collision-Free Minimum-Jerk Robot Motion Generation},
  author={Sundaralingam, Balakumar and Hari, Siva Kumar Sastry and Fishman, Adam and Garrett, Caelan and Van Wyk, Karl and Blukis, Valts and Millane, Alexander and Oleynikova, Helen and Handa, Ankur and Ramos, Fabio and Ratliff, Nathan and Fox, Dieter},
  journal={arXiv preprint arXiv:2310.17274},
  year={2023}
}

@inproceedings{lpips,
  title={The unreasonable effectiveness of deep features as a perceptual metric},
  author={Zhang, Richard and Isola, Phillip and Efros, Alexei A and Shechtman, Eli and Wang, Oliver},
  booktitle=CVPR,
  year={2018}
}

@misc{lightx2v,
  author={LightX2V Contributors},
  title={LightX2V: Light Video Generation Inference Framework},
  year={2025},
  publisher={GitHub},
  url={https://github.com/ModelTC/lightx2v}
}

@inproceedings{planet,
  title={Learning latent dynamics for planning from pixels},
  author={Hafner, Danijar and Lillicrap, Timothy and Fischer, Ian and Villegas, Ruben and Ha, David and Lee, Honglak and Davidson, James},
  booktitle=ICML,
  year={2019}
}

@article{dreamerv3,
  title={Mastering diverse domains through world models},
  author={Hafner, Danijar and Pasukonis, Jurgis and Ba, Jimmy and Lillicrap, Timothy},
  journal={arXiv preprint arXiv:2301.04104},
  year={2023}
}

@inproceedings{daydreamer,
  title={Daydreamer: World models for physical robot learning},
  author={Wu, Philipp and Escontrela, Alejandro and Hafner, Danijar and Abbeel, Pieter and Goldberg, Ken},
  booktitle=CORL,
  year={2023}
}

@inproceedings{td-mpc2,
  title={Td-mpc2: Scalable, robust world models for continuous control},
  author={Hansen, Nick and Su, Hao and Wang, Xiaolong},
  booktitle=ICLR,
  year={2024}
}

@inproceedings{genie,
  title={Genie: Generative interactive environments},
  author={Bruce, Jake and Dennis, Michael D and Edwards, Ashley and Parker-Holder, Jack and Shi, Yuge and Hughes, Edward and Lai, Matthew and Mavalankar, Aditi and Steigerwald, Richie and Apps, Chris and others},
  booktitle=ICML,
  year={2024}
}

@article{diamond,
  title={Diffusion for world modeling: Visual details matter in atari},
  author={Alonso, Eloi and Jelley, Adam and Micheli, Vincent and Kanervisto, Anssi and Storkey, Amos and Pearce, Tim and Fleuret, Fran{\c{c}}ois},
  journal=NIPS,
  year={2024}
}

@inproceedings{gamengen,
  title={Diffusion models are real-time game engines},
  author={Valevski, Dani and Leviathan, Yaniv and Arar, Moab and Fruchter, Shlomi},
  booktitle=ICLR,
  year={2025}
}

@inproceedings{unisim,
  title={Unisim: A neural closed-loop sensor simulator},
  author={Yang, Ze and Chen, Yun and Wang, Jingkang and Manivasagam, Sivabalan and Ma, Wei-Chiu and Yang, Anqi Joyce and Urtasun, Raquel},
  booktitle=CVPR,
  year={2023}
}

@article{ivideogpt,
  title={ivideogpt: Interactive videogpts are scalable world models},
  author={Wu, Jialong and Yin, Shaofeng and Feng, Ningya and He, Xu and Li, Dong and Hao, Jianye and Long, Mingsheng},
  journal=NIPS,
  year={2024}
}

@inproceedings{nwm,
  title={Navigation world models},
  author={Bar, Amir and Zhou, Gaoyue and Tran, Danny and Darrell, Trevor and LeCun, Yann},
  booktitle=CVPR,
  year={2025}
}

@inproceedings{aether,
  title={Aether: Geometric-aware unified world modeling},
  author={Zhu, Haoyi and Wang, Yifan and Zhou, Jianjun and Chang, Wenzheng and Zhou, Yang and Li, Zizun and Chen, Junyi and Shen, Chunhua and Pang, Jiangmiao and He, Tong},
  booktitle=ICCV,
  year={2025}
}

@inproceedings{seva,
  title={Stable virtual camera: Generative view synthesis with diffusion models},
  author={Zhou, Jensen and Gao, Hang and Voleti, Vikram and Vasishta, Aaryaman and Yao, Chun-Han and Boss, Mark and Torr, Philip and Rupprecht, Christian and Jampani, Varun},
  booktitle=ICCV,
  year={2025}
}

@article{gaia,
  title={Gaia-2: A controllable multi-view generative world model for autonomous driving},
  author={Russell, Lloyd and Hu, Anthony and Bertoni, Lorenzo and Fedoseev, George and Shotton, Jamie and Arani, Elahe and Corrado, Gianluca},
  journal={arXiv preprint arXiv:2503.20523},
  year={2025}
}

@article{adaworld,
  title={Adaworld: Learning adaptable world models with latent actions},
  author={Gao, Shenyuan and Zhou, Siyuan and Du, Yilun and Zhang, Jun and Gan, Chuang},
  journal={arXiv preprint arXiv:2503.18938},
  year={2025}
}

@article{coworld,
  title={CoWorld-VLA: Thinking in a Multi-Expert World Model for Autonomous Driving},
  author={Huang, Minqing and Xiang, Yujiao and Liang, Zihan and Huang, Jiajie and Wang, Jingqi and Xu, Zhi and Tan, Feiyang and Zhou, Hangning and Yang, Mu and Che, Gong},
  journal={arXiv preprint arXiv:2605.10426},
  year={2026}
}

@article{clam,
  title={Clam: Continuous latent action models for robot learning from unlabeled demonstrations},
  author={Liang, Anthony and Czempin, Pavel and Hong, Matthew and Zhou, Yutai and Biyik, Erdem and Tu, Stephen},
  journal={arXiv preprint arXiv:2505.04999},
  year={2025}
}

@article{univla,
  title={Univla: Learning to act anywhere with task-centric latent actions},
  author={Bu, Qingwen and Yang, Yanting and Cai, Jisong and Gao, Shenyuan and Ren, Guanghui and Yao, Maoqing and Luo, Ping and Li, Hongyang},
  journal={arXiv preprint arXiv:2505.06111},
  year={2025}
}

@article{lawm,
  title={Learning Latent Action World Models In The Wild},
  author={Garrido, Quentin and Nagarajan, Tushar and Terver, Basile and Ballas, Nicolas and LeCun, Yann and Rabbat, Michael},
  journal={arXiv preprint arXiv:2601.05230},
  year={2026}
}

@inproceedings{controlnet,
  title={Adding conditional control to text-to-image diffusion models},
  author={Zhang, Lvmin and Rao, Anyi and Agrawala, Maneesh},
  booktitle=ICCV,
  year={2023}
}

@inproceedings{controlvideo,
  title={Controlvideo: Training-free controllable text-to-video generation},
  author={Zhang, Yabo and Wei, Yuxiang and ZHANG, XIAOPENG and Zuo, Wangmeng and Tian, Qi and others},
  booktitle=ICLR,
  year={2024}
}

@article{videocomposer,
  title={Videocomposer: Compositional video synthesis with motion controllability},
  author={Wang, Xiang and Yuan, Hangjie and Zhang, Shiwei and Chen, Dayou and Wang, Jiuniu and Zhang, Yingya and Shen, Yujun and Zhao, Deli and Zhou, Jingren},
  journal=NIPS,
  year={2023}
}

@article{animatediff,
  title={Animatediff: Animate your personalized text-to-image diffusion models without specific tuning},
  author={Guo, Yuwei and Yang, Ceyuan and Rao, Anyi and Liang, Zhengyang and Wang, Yaohui and Qiao, Yu and Agrawala, Maneesh and Lin, Dahua and Dai, Bo},
  journal={arXiv preprint arXiv:2307.04725},
  year={2023}
}

@inproceedings{motionctrl,
  title={Motionctrl: A unified and flexible motion controller for video generation},
  author={Wang, Zhouxia and Yuan, Ziyang and Wang, Xintao and Li, Yaowei and Chen, Tianshui and Xia, Menghan and Luo, Ping and Shan, Ying},
  booktitle=SIGGRAPH,
  year={2024}
}

@article{dragnuwa,
  title={Dragnuwa: Fine-grained control in video generation by integrating text, image, and trajectory},
  author={Yin, Shengming and Wu, Chenfei and Liang, Jian and Shi, Jie and Li, Houqiang and Ming, Gong and Duan, Nan},
  journal={arXiv preprint arXiv:2308.08089},
  year={2023}
}

@inproceedings{tora,
  title={Tora: Trajectory-oriented diffusion transformer for video generation},
  author={Zhang, Zhenghao and Liao, Junchao and Li, Menghao and Dai, Zuozhuo and Qiu, Bingxue and Zhu, Siyu and Qin, Long and Wang, Weizhi},
  booktitle=CVPR,
  year={2025}
}

@inproceedings{mofa,
  title={Mofa-video: Controllable image animation via generative motion field adaptions in frozen image-to-video diffusion model},
  author={Niu, Muyao and Cun, Xiaodong and Wang, Xintao and Zhang, Yong and Shan, Ying and Zheng, Yinqiang},
  booktitle=ECCV,
  year={2024}
}

@inproceedings{trackgo,
  title={Trackgo: A flexible and efficient method for controllable video generation},
  author={Zhou, Haitao and Wang, Chuang and Nie, Rui and Liu, Jinlin and Yu, Dongdong and Yu, Qian and Wang, Changhu},
  booktitle=AAAI,
  year={2025}
}

@inproceedings{vace,
  title={Vace: All-in-one video creation and editing},
  author={Jiang, Zeyinzi and Han, Zhen and Mao, Chaojie and Zhang, Jingfeng and Pan, Yulin and Liu, Yu},
  booktitle=ICCV,
  year={2025}
}

@inproceedings{motion-prompting,
  title={Motion prompting: Controlling video generation with motion trajectories},
  author={Geng, Daniel and Herrmann, Charles and Hur, Junhwa and Cole, Forrester and Zhang, Serena and Pfaff, Tobias and Lopez-Guevara, Tatiana and Aytar, Yusuf and Rubinstein, Michael and Sun, Chen and others},
  booktitle=CVPR,
  year={2025}
}

@article{motionstream,
  title={Motionstream: Real-time video generation with interactive motion controls},
  author={Shin, Joonghyuk and Li, Zhengqi and Zhang, Richard and Zhu, Jun-Yan and Park, Jaesik and Shechtman, Eli and Huang, Xun},
  journal={arXiv preprint arXiv:2511.01266},
  year={2025}
}

@article{freetraj,
  title={Freetraj: Tuning-free trajectory control in video diffusion models},
  author={Qiu, Haonan and Chen, Zhaoxi and Wang, Zhouxia and He, Yingqing and Xia, Menghan and Liu, Ziwei},
  journal={arXiv preprint arXiv:2406.16863},
  year={2024}
}

@inproceedings{trailblazer,
  title={Trailblazer: Trajectory control for diffusion-based video generation},
  author={Ma, Wan-Duo Kurt and Lewis, John P and Kleijn, W Bastiaan},
  booktitle=SIGGRAPHASIA,
  year={2024}
}

@inproceedings{peekaboo,
  title={Peekaboo: Interactive video generation via masked-diffusion},
  author={Jain, Yash and Nasery, Anshul and Vineet, Vibhav and Behl, Harkirat},
  booktitle=CVPR,
  year={2024}
}

@inproceedings{image-conductor,
  title={Image conductor: Precision control for interactive video synthesis},
  author={Li, Yaowei and Wang, Xintao and Zhang, Zhaoyang and Wang, Zhouxia and Yuan, Ziyang and Xie, Liangbin and Shan, Ying and Zou, Yuexian},
  booktitle=AAAI,
  year={2025}
}

@article{minimaxremover,
  title={Minimax-remover: Taming bad noise helps video object removal},
  author={Zi, Bojia and Peng, Weixuan and Qi, Xianbiao and Wang, Jianan and Zhao, Shihao and Xiao, Rong and Wong, Kam-Fai},
  journal={Advances in Neural Information Processing Systems},
  volume={38},
  pages={75518--75547},
  year={2026}
}

\clearpage
\appendix
\section{Interface Construction Details}
\label{sec:supp:interface}

\paragraph{Dataset-specific nominalization.}
The main paper distinguishes raw actions, nominal robot-only trajectories, and realized robot states.
For DROID~\cite{droid}, demonstrations are collected through human teleoperation, and the dataset provides standardized robot action representations, including joint-space and end-effector-space actions.
In our preprocessing, we use the logged joint and gripper target sequence as the raw action for nominal replay.
These targets can be specified faster than the arm can physically track; directly rendering them therefore often produces motion that runs ahead of the recorded video.
We replay the action in a scene-free, robot-only Isaac Lab~\cite{mittal2025isaaclab} environment, where the Panda controller and actuation limits shape the raw action sequence into a physically trackable nominal trajectory.

For RoboCasa-GR1, the DiT4DiT~\cite{dit4dit} policy outputs 29D controller actions, while the renderer consumes the 39D Fourier GR-1 joint/hand/waist state.
We replay each action sequence from the clip-start state in a collision-free shadow rollout and render the resulting Fourier GR-1 nominal state.
The RoboCasa-GR1 corpus consists of episode trajectories generated by executing a DiT4DiT VLA policy with the Fourier GR-1 humanoid tabletop setup released by the GR00T~\cite{gr00t} project on top of the RoboCasa~\cite{robocasa} simulator.\footnote{\url{https://github.com/robocasa/robocasa-gr1-tabletop-tasks}}
In both robot datasets, the world model receives nominal robot geometry as its deployment-compatible conditioning signal.

\paragraph{Static context and scene descriptions.}
The static context stream follows the camera trajectory of each target clip while removing robot interaction.
For fixed-view DROID clips, the RGB context is the initial observation repeated across the prediction horizon, including the robot if visible; it is not a robot-free reconstruction.
When scene depth is enabled, we use metric depth estimated from the ZED stereo observations with FoundationStereo~\cite{wen2025stereo} and align it to the same fixed camera frame.
For RoboCasa-GR1 dynamic-view clips, the simulator provides the camera trajectory and robot-free scene, so we render both static RGB and static depth along that trajectory without the robot.
This construction supplies camera-aware scene context while leaving action information in the nominal robot mesh and end-effector depth streams.

Each clip is paired with a scene description $\mathcal{T}$ generated by Qwen3-VL~\cite{qwen3vl}.
This language condition describes visible scene appearance and context only.
It provides scene-level semantic context while the nominal robot geometry carries action information.

\section{Model, Baseline, and Evaluation Details}
\label{sec:supp:training}

\paragraph{Wan adaptation.}
The Wan~\cite{wan2025} 2.1 14B variants are built on the pretrained Wan2.1-Fun-V1.1-14B-InP image-to-video model from VideoX-Fun~\cite{videox-fun}.
We adapt it with two trainable additions while freezing all other pretrained weights:
(i) the patch-embedding projection is extended to ingest the extra conditioning channels, and
(ii) LoRA~\cite{hu2022lora} weights with rank~64 and $\alpha{=}64$ are attached to the transformer.
The base model has 36 input latent channels.
Each rendered stream is VAE-encoded to a 16-channel latent and concatenated along the channel axis, so the patch embedding grows to $36+C$ input channels, where $C$ is the number of conditioning channels in the variant.
The extended patch-embedding weights are trained together with the LoRA weights.
Training uses the latent flow-matching objective~\cite{lipman2023flowmatching,liu2023flow}: noisy latents are linearly interpolated between data and noise, and the DiT predicts the corresponding velocity.
For the nominal/logged-state diagnostic, the DROID nominal model is trained for 15,010 steps and the remaining models for 10,000 steps.

\paragraph{Conditioning channels.}
Each rendered stream contributes 16 latent channels:
\begin{itemize}
  \item \emph{Mesh RGB only} (raw action mesh or nominal trajectory rendering): $C=16$.
  \item \emph{Mesh RGB $+$ end-effector depth $+$ scene depth}: $C=48$.
\end{itemize}
The static RGB context follows the Wan2.1-Fun InP image-conditioning path, and the extra rendered-stream path carries robot mesh and depth conditions.
The static video contributes 16 latent channels, and the inpainting mask contributes four time-compressed mask channels.
We set the mask to all-known pixels ($m{=}1$ in our convention), so the static stream preserves scene appearance while the rendered robot streams explain action-conditioned changes.
Inspired by Dexterous World Models~\cite{dwm}, this separates static scene conditioning from action-conditioned interaction dynamics.

\paragraph{AdaLN state-vector baseline.}
We use AdaLN~\cite{peebles2023scalable} conditioning to inject numeric state vectors into the diffusion transformer.
For DROID, the AdaLN baseline uses a per-frame 7D nominal end-effector state,
$[x,y,z,\mathrm{roll},\mathrm{pitch},\mathrm{yaw},g]$,
computed by Panda forward kinematics from the robot-only Isaac Lab nominal joint rollout and the nominal gripper scalar $g$.
For RoboCasa-GR1, the DiT4DiT~\cite{dit4dit} policy action is a 29D controller action, ordered as left arm 7, right arm 7, left hand 6, right hand 6, and waist 3.
After replay from the clip-start state, the AdaLN baseline uses the resulting 39D nominal Fourier GR-1 state, ordered as left arm 7, right arm 7, left hand 11, right hand 11, and waist 3.
For joint DROID+RoboCasa training, DROID fills the first 7 entries and RoboCasa fills the next 39 entries of a zero-padded 46D union vector.
Four raw-frame states are stacked for each Wan latent frame, giving a 184D AdaLN input.

\begin{table}[t]
\centering
\small
\caption{Training hyperparameters for the Wan 2.1 14B world model.}
\label{tab:supp:hparams}
\begin{tabular}{ll}
\toprule
Hyperparameter & Value \\
\midrule
Base model              & Wan2.1-Fun-V1.1-14B-InP \\
LoRA rank / $\alpha$    & 64 / 64 \\
Trainable parameters    & extended patch embedding $+$ LoRA \\
Optimizer               & AdamW~\cite{loshchilov2018decoupled} ($\beta_1{=}0.9$, $\beta_2{=}0.95$) \\
Weight decay            & 0.01 \\
Learning rate           & $1\times10^{-4}$ \\
LR schedule             & cosine with restarts, 100 warmup steps \\
Gradient clipping       & 1.0 \\
Batch size / grad.\ accum.\ & 1 / 4 per GPU \\
Precision               & bf16 \\
Gradient checkpointing  & enabled \\
Resolution              & $480\times832$ \\
Clip length / fps       & 81 frames / 16 \\
\bottomrule
\end{tabular}
\end{table}

\paragraph{Inference.}
Wan 2.1 14B variants are sampled with the LightX2V~\cite{lightx2v} CFG-and-step-distilled LoRA at 4 denoising steps and guidance scale 1.0.
SVD-based variants use 25 Euler steps at $192{\times}320$.
Because we extend the pretrained Wan 2.1 14B transformer through a patch-embedding projection and task-specific LoRA weights, the LightX2V distillation LoRA applies directly and reduces denoising steps by $10{\times}$.

\section{Robot-Excluded Evaluation}
\label{sec:supp:masked}

We assess how much of the reconstruction advantage remains outside the robot region.
Masks are obtained from camera-aligned renders of the ground-truth robot state and shared between methods.
PSNR pools squared errors over non-robot pixels.
SSIM averages the local score map over the non-robot region after a five-pixel erosion to avoid windows that cross the robot boundary.
For LPIPS, we replace the predicted robot region with ground-truth pixels before computing the perceptual distance.
This suppresses direct robot reconstruction errors, but feature receptive fields mean it is not a strictly pixel-local object metric.
The mask threshold is 16 on the 8-bit render, with no robot-mask dilation.

\begin{table}[t]
\centering
\small
\setlength{\tabcolsep}{4pt}
\caption{\textbf{Full-frame and robot-excluded evaluation on matched clips.}
RoboCasa-GR1 includes only the 82 clips with available ground-truth robot masks.
Robot-excluded LPIPS uses ground-truth filling as defined in the text.}
\label{tab:supp:masked}
\begin{tabular}{llcccccc}
\toprule
& & \multicolumn{3}{c}{Full frame} & \multicolumn{3}{c}{Robot excluded} \\
\cmidrule(lr){3-5}\cmidrule(lr){6-8}
Dataset & Method & PSNR$\uparrow$ & SSIM$\uparrow$ & LPIPS$\downarrow$ & PSNR$\uparrow$ & SSIM$\uparrow$ & LPIPS$\downarrow$ \\
\midrule
DROID & AdaLN & 18.57 & 0.824 & 0.224 & 20.87 & 0.872 & 0.176 \\
($n{=}256$) & Ours & 21.87 & 0.859 & 0.178 & 23.71 & 0.893 & 0.150 \\
\midrule
RoboCasa-GR1 & AdaLN & 18.05 & 0.844 & 0.184 & 20.85 & 0.893 & 0.141 \\
($n{=}82$) & Ours & 24.89 & 0.892 & 0.129 & 26.42 & 0.914 & 0.117 \\
\bottomrule
\end{tabular}
\end{table}

Evaluation covers all 256 DROID test clips and the 82 of 128 RoboCasa-GR1 test clips with available ground-truth robot masks.
Within each dataset, the full-frame and robot-excluded columns in Table~\ref{tab:supp:masked} use identical clips and predictions.
For AdaLN on RoboCasa-GR1, predictions are sampled in a separate evaluation run using the same checkpoint and static-context configuration as the main comparison.
The advantage outside the robot region supports improved non-robot reconstruction.

\section{Embodiment Transfer Protocols}
\label{sec:supp:transfer}

\paragraph{RoboTwin single-arm transfer.}
We evaluate ur5-wsg and ARX-X5 with eight tasks and five episodes per task, using the joint DROID+RoboCasa-GR1 checkpoint without target-robot training.
Both interfaces are constructed from the same nominal motions.
AdaLN fills its existing pose--gripper slot and requires no input-layer modification in this compatible single-arm setting.
Ours changes the rendered URDF geometry while retaining the trained conditioning path.
The transfer table reports full-frame metrics.
These tests span changes in both embodiment and environment.

\paragraph{HRDexDB dexterous adaptation.}
The object-disjoint split contains 575 training clips from 192 episodes and 30 objects, and 84 evaluation clips from 84 episodes and ten other objects.
Clips are 81 frames at $480{\times}640$, anchored around the grasp, with training offsets of 0, $-2$, and $-4$ seconds.
We evaluate at this native aspect ratio.
Both methods are fine-tuned for the same number of steps with the same learning rate, schedule, accumulation, and seed.
AdaLN uses a 12D input comprising end-effector translation, rotation vector, and six commanded hand joints, with four-frame delta stacking into 48D.
The input projection and state-normalization parameters are reinitialized; remaining compatible weights are reused.
Ours keeps its mesh/depth interface for both zero-shot and fine-tuned evaluation.
The fine-tuned comparison uses equal optimization budgets with these method-specific input adaptations.

\section{Task-Outcome Evaluation Protocols}
\label{sec:supp:outcomes}

\paragraph{Judge and labels.}
We sample 12 frames uniformly from each generated video and query GPT-5.4 once with a task-specific success criterion, requesting a structured binary prediction.
The same judge configuration is used for both world models.
Ground-truth labels are simulator success predicates.
Task descriptions are supplied to the evaluator; world-model language conditioning remains the scene-only description described in Appendix~\ref{sec:supp:interface}.
We report binary classification accuracy against simulator outcomes, reflecting both video-generation quality and judge reliability.

\paragraph{RoboCasa-GR1 selection.}
The evaluation pool contains 256 held-out clips over 22 tasks, separate from the reconstruction evaluation set.
The judge correctly recovers the simulator labels from ground-truth video for 209 clips within the prediction window.
From these clips, we select 80 success/failure pairs within tasks, retaining all 22 tasks.
Within-task selection is lexicographic by episode identifier, and excess pairs are removed from the largest task groups until 80 pairs remain.
Selection is conditioned on ground-truth-video observability and is independent of either model's generated-video predictions.
On the complete 256-clip set, accuracy is 64.5\% for AdaLN and 66.4\% for Ours; the same judge reaches 81.6\% on ground-truth video.

\paragraph{RoboLab protocol.}
We collect policy rollouts on BowlStackingLeftOnRight, ReorientWhiteMugs, CubesAndBlocksInBin, MustardInRightBin, and PickGlasses, selecting ten simulator successes and ten failures per task.
Both methods predict 101 frames: an 81-frame action sequence followed by 20 frames holding the final pose, so terminal outcomes have time to become visible.
For the 101-frame setting, the success-positive confusion counts (TP, TN, FP, FN) are $(0,49,1,50)$ for AdaLN and $(18,48,2,32)$ for Ours.
Both are conservative about predicting success, especially AdaLN; substantial false-negative error remains even for Ours.
Each example consists of one recorded action sequence and one predicted rollout.

\section{Human Demonstration to Robot Video Pipeline}
\label{sec:supp:h2r-pipeline}

\begin{figure}[t]
  \centering
  \includegraphics[width=\linewidth]{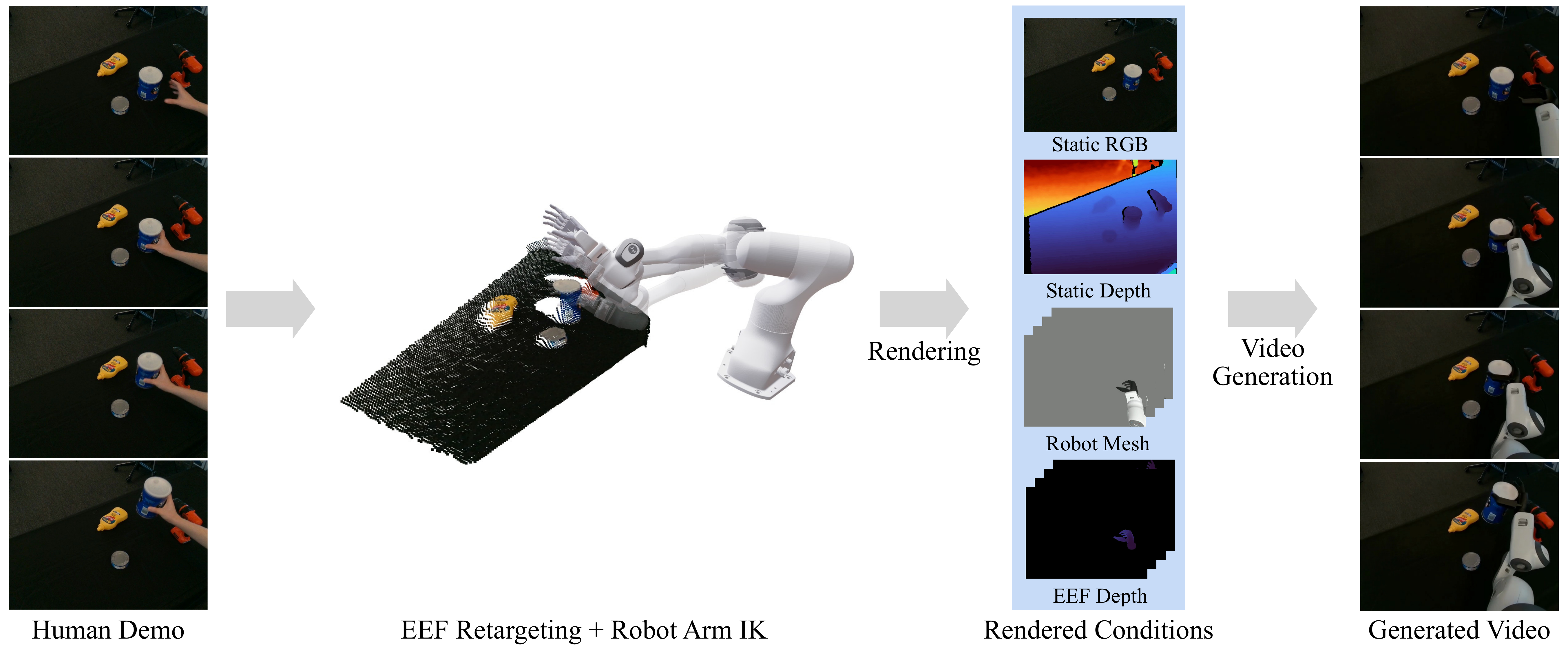}
  \caption{\textbf{Human demonstration to robot video pipeline.}
  Human motion is converted into the same rendered mesh-and-depth interface used for robot data before being passed to the world model.}
  \label{fig:supp:h2r-pipeline}
\end{figure}

Figure~\ref{fig:supp:h2r-pipeline} illustrates how a DexYCB~\cite{dexycb} RGB-D clip with per-frame 21-point 3D hand keypoints is converted into the same rendered robot interface used for robot demonstrations.

\emph{Scene inpainting.}
When the initial frame contains a visible human hand, we optionally remove it with a video inpainting model~\cite{minimaxremover}.
This yields a robot-free scene image for the static RGB context; clips without a visible hand in the initial frame use the original scene observation.

\emph{Hand retargeting.}
We use the AnyTeleop~\cite{qin2023anyteleop} retargeting setup to map human fingertip motion to the robot hand.

\emph{Arm retargeting.}
The wrist trajectory and hand orientation define a target end-effector motion, which is mapped to the robot arm with PyRoki~\cite{kim2025pyroki} inverse kinematics.

\emph{Mesh rendering and conditioning assembly.}
The combined Panda arm and Inspire hand trajectory is rendered in the DexYCB camera frame using the calibrated robot geometry.
We then assemble the same conditioning streams used by the robot world model: static RGB, static depth, robot mesh video, and end-effector depth.
The same trained world model receives these streams and generates a robot manipulation rollout from the human-sourced motion.

\section{Additional Qualitative Results}
\label{sec:supp:qualitative}

\subsection{Counterfactual Trajectory Edits}
\label{sec:supp:counterfactual}

\begin{figure}[t]
  \centering
  \includegraphics[width=\linewidth]{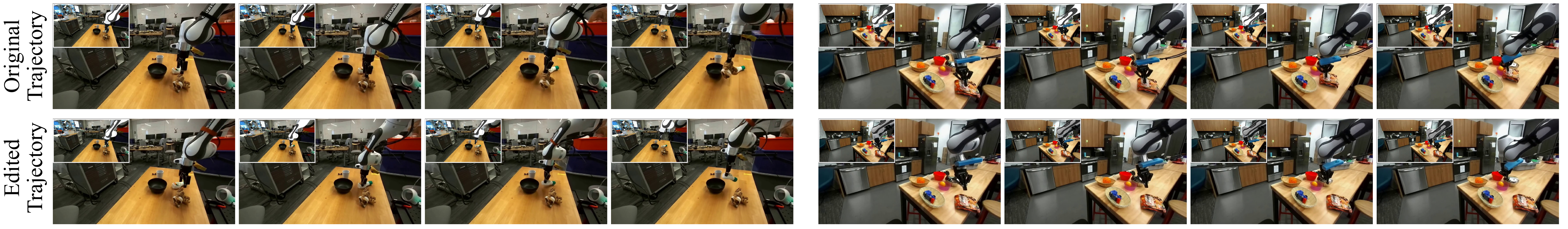}
  \caption{\textbf{Response to trajectory edits.} With the initial scene fixed, editing the rendered nominal trajectory changes the predicted scene response.}
  \label{fig:counterfactual}
\end{figure}

On DROID episodes, we keep the initial scene fixed and compare the original nominal rendering with an edited trajectory re-solved by cuRobo~\cite{curobo} under the same Panda kinematics.
Changing only the rendered nominal trajectory changes the predicted scene response (Figure~\ref{fig:counterfactual}).
The edited trajectories have no corresponding executed ground-truth videos, so this probe assesses the model's qualitative response to trajectory edits.

\subsection{Additional Embodiments and Comparisons}
\label{sec:supp:embodiments}

\begin{figure}[p]
  \centering
  \includegraphics[width=\linewidth]{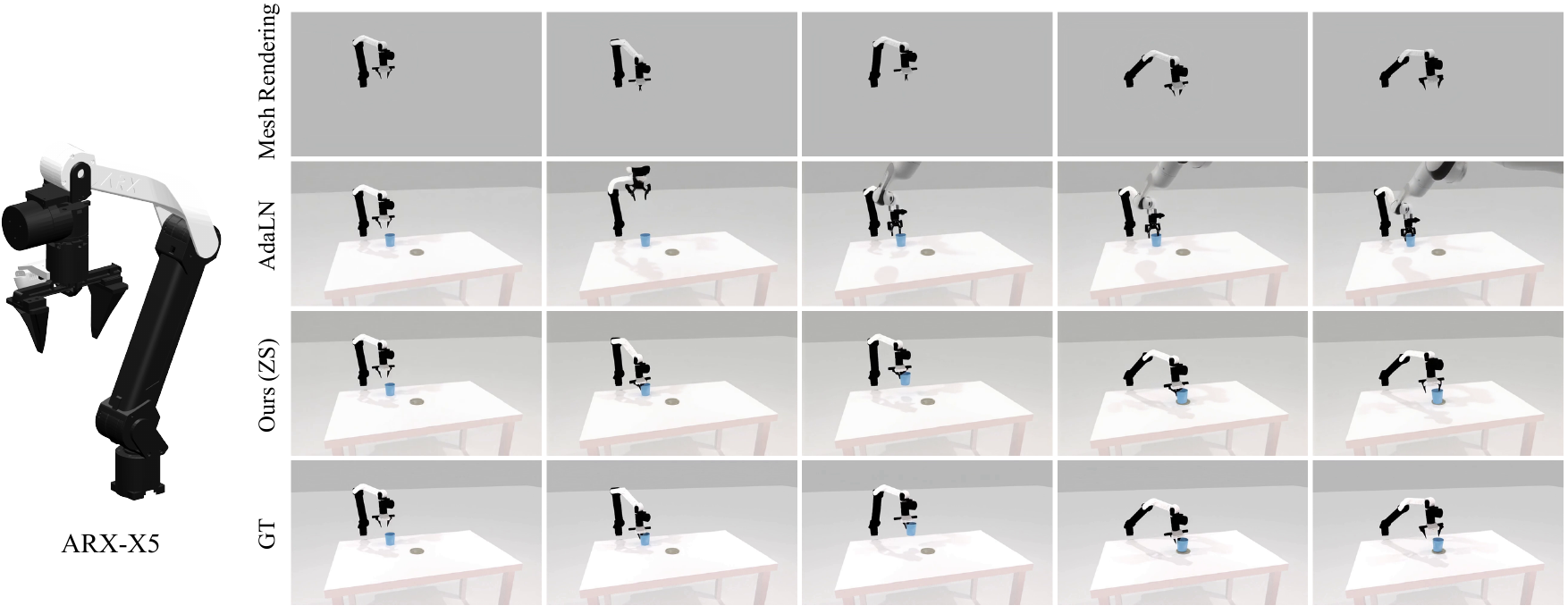}\\[1em]
  \includegraphics[width=\linewidth]{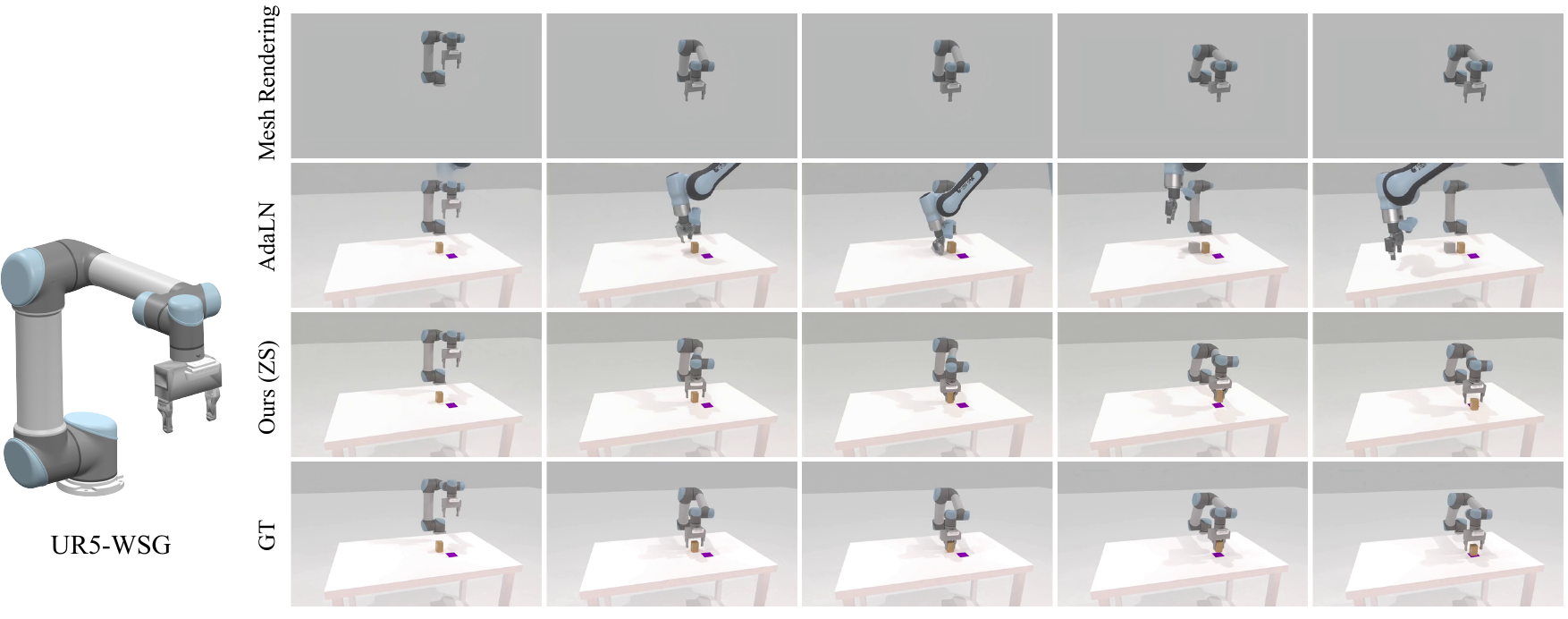}
  \caption{\textbf{Zero-shot transfer on RoboTwin.}
  ARX-X5 cup placement (top) and UR5--WSG stamp sealing (bottom).
  Each panel shows the embodiment at left and, from top to bottom, the rendered nominal trajectory, AdaLN and Ours predictions, and ground-truth video.
  Both methods are evaluated zero-shot.}
  \label{fig:supp:robotwin-zero-shot}
\end{figure}

\paragraph{Zero-shot single-arm transfer.}
Figure~\ref{fig:supp:robotwin-zero-shot} compares AdaLN and Ours on ARX-X5 cup placement and UR5--WSG stamp sealing from RoboTwin.
Both methods use the joint DROID+RoboCasa-GR1 checkpoint without target-robot fine-tuning, following the protocol in Appendix~\ref{sec:supp:transfer}.

\paragraph{Zero-shot multi-embodiment inference.}
\begin{figure}[tp]
  \centering
  \includegraphics[width=\linewidth]{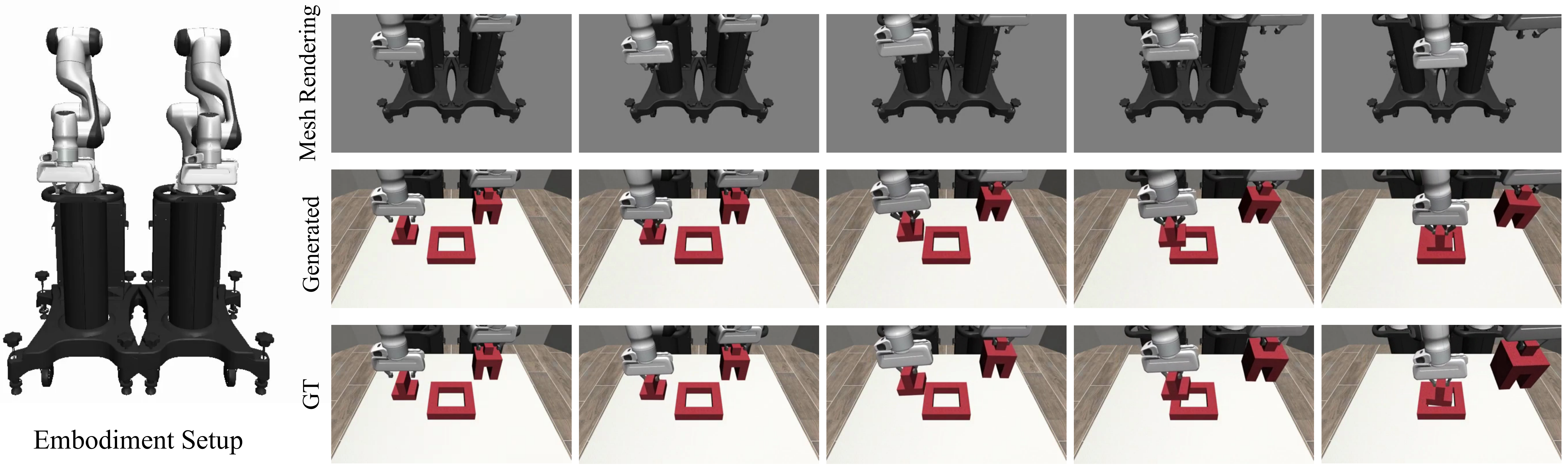}
  \caption{\textbf{Zero-shot multi-embodiment inference.}
  A bimanual Franka Panda setup from DexMimicGen is rendered as robot geometry and passed to the same video model.}
  \label{fig:supp:dexmimicgen-zero-shot}
\end{figure}
Figure~\ref{fig:supp:dexmimicgen-zero-shot} probes zero-shot inference on a DexMimicGen~\cite{jiang2025dexmimicgen} scenario with two Franka Panda arms and parallel-jaw grippers.
Because the conditioning signal is rendered robot geometry in the camera frame, the same video model can consume the new multi-arm prompt through the rendered interface.
This qualitative result illustrates how the rendered interface supports new embodiment and end-effector configurations through the same conditioning path.

\begin{figure}[p]
  \centering
  \includegraphics[width=\linewidth,height=0.88\textheight,keepaspectratio]{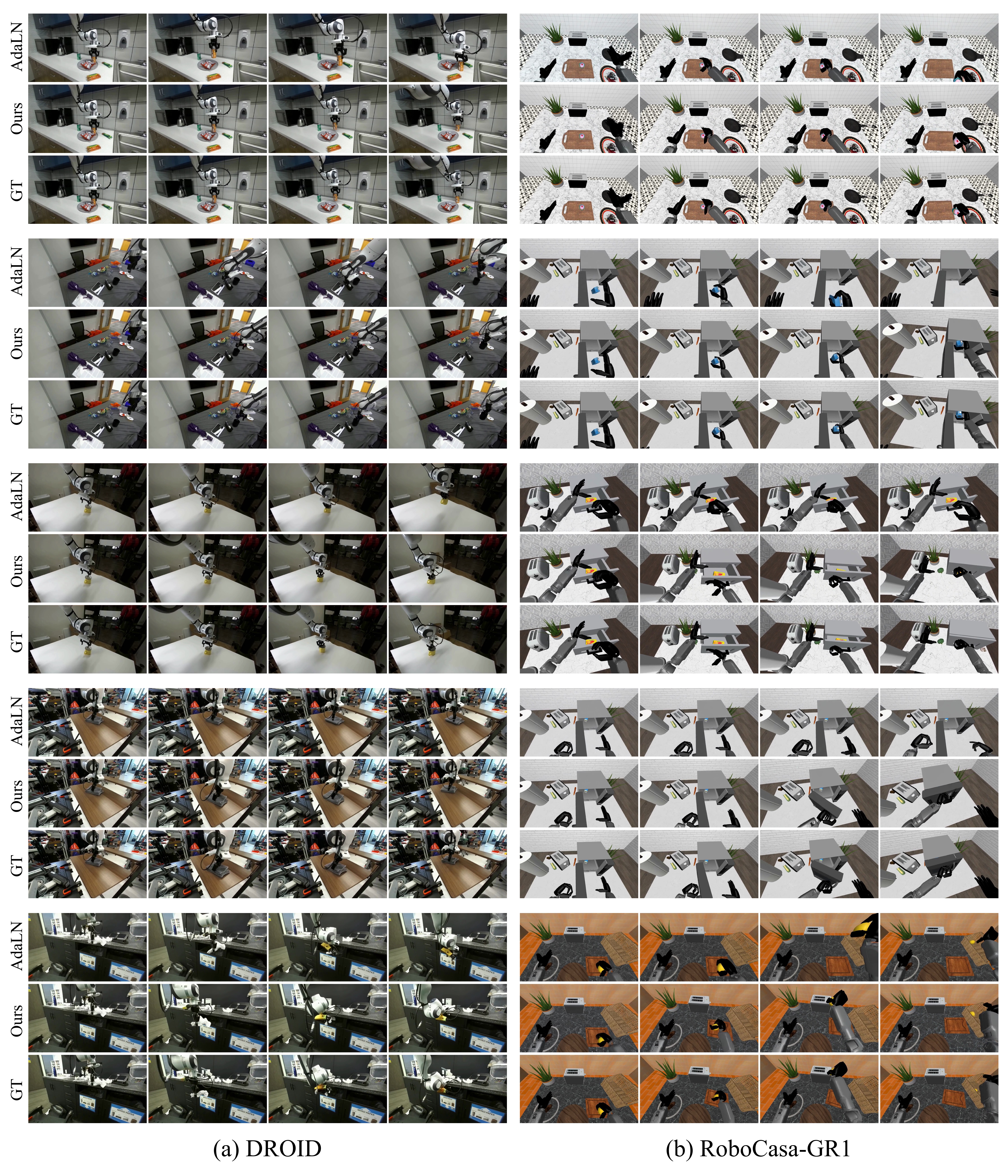}
  \caption{\textbf{Additional qualitative comparisons on DROID and RoboCasa-GR1.}
  Rendered robot geometry exposes the action in the target camera frame and more consistently follows the commanded robot motion than AdaLN state-vector conditioning.}
  \label{fig:supp:qualitative-comparisons}
\end{figure}

\paragraph{DROID and RoboCasa-GR1 comparisons.}
Figure~\ref{fig:supp:qualitative-comparisons} provides additional qualitative comparisons, with DROID examples on the left and RoboCasa-GR1 examples on the right.
Across these examples, AdaLN state-vector conditioning can preserve scene appearance but often weakly realizes or misplaces the intended robot motion.
Rendered robot geometry makes the action visible in the target camera frame, so the generated videos more consistently follow the commanded motion and localize scene changes around the robot and contact region.

\end{document}